\documentclass[twoside]{article}
\usepackage{xcolor}

\usepackage[accepted]{aistats2023}
\usepackage[round]{natbib}

\usepackage{mathtools}
\usepackage{amsthm}
\usepackage{amsfonts}
\usepackage{amsmath}
\usepackage{amssymb}
\usepackage{bbm}
\usepackage{booktabs} 
\usepackage{graphicx}
\usepackage{subfigure}
\usepackage{multirow}
\usepackage{color}
\usepackage{tikz}
\graphicspath{{imgs/}}
\newtheorem{theorem}{Theorem}[section]
\newtheorem{corollary}{Corollary}[theorem]
\newtheorem{lemma}{Lemma}[section]

\theoremstyle{definition}
\newtheorem{definition}{Definition}[section]

\newcommand{\REF}[1]{(\ref{#1})}

\newcommand{\argmin}{\operatornamewithlimits{argmin}}

\newcommand{\fl}{\operatorname{fl}}
\newcommand{\op}{\operatorname{op}}

\def\ZZ{\mathbb{Z}}
\def\NN{\mathbb{N}}

\def\BB{\mathbb{B}}
\def\RR{{\rm I\!R}}
\def\NN{\mathbb{N}}

\def\FF{\mathbb{F}}

\def\S{{\cal S}}

\def\w{\mathbf{w}}

\def\V{{\cal V}}

\def\S{{\cal S}}
\def\s{{\cal s}}
\def\F{{\cal F}}

\def\d{\delta}

\DeclareMathOperator*{\sgn}{sgn}
\DeclareMathOperator*{\dist}{dist}

\def\EE{{\rm I\! E}}
\def\n{d} %dimensnion of space $n$
\def\RRn{\RR^\n} % domain of weights \RR^n
\def\x{\mathbf w} % optimization parameter $x$
\def\Xs{\mathbf \Omega^*} %optimal point $X^*$
\def\X{\mathbf \Omega} %optimization set 
\def \piX{\mathsf P_\Omega}
\def\g{\mathbf g}
\def\s{\mathbf s}
\def\r{\mathbf r}
\def\y{\mathbf y}

\def\t{\eta} %learning rate
\newcommand{\norm}[1]{\left\lVert#1\right\rVert} %norm

\def\V{\n\sigma^2_r}
\def\W{\n\sigma^2_s}

\begin{document}



\twocolumn[

\aistatstitle{On the Convergence of Stochastic Gradient Descent in Low-precision Number Formats}

\aistatsauthor{ Matteo Cacciola$^1$ \quad Antonio Frangioni$^2$ \quad  Masoud Asgharian$^3$ \quad  Alireza Ghaffari$^4$ \quad Vahid Partovi Nia$^4$ }

\aistatsaddress{ $^1$Polytechnique Montreal \quad  $^2$University of Pisa \quad $^3$McGill University \quad $^4$Huawei Noah's Ark Lab } ]

\begin{abstract}
Deep learning models are dominating almost all artificial intelligence tasks such as vision, text, and speech processing. Stochastic Gradient Descent (SGD) is the main tool for training such models, where the computations are usually performed in single-precision floating-point number format. The convergence of single-precision SGD is normally aligned with the theoretical results of real numbers since they exhibit negligible error. However, the numerical error increases when the computations are performed in low-precision number formats. This provides compelling reasons to study the SGD convergence adapted for low-precision computations. We present both deterministic and stochastic analysis of the SGD algorithm, obtaining bounds that show the effect of number format. Such bounds can provide guidelines as to how SGD convergence is affected when constraints render the possibility of performing high-precision computations remote.
%Thus, here we present an analysis of SGD convergence under plausible assumptions, supported by our numerical experiments.

% The convergence of Artificial Neural Networks (ANNs) training have been studied under different hypothesis. However, is rarely taken into account the fact that in practice this models are trained using computers that, obviously, have finite computation precision. This may seem not relevant since the machine error is very small, but this is starting to not be the case for ANNs training. There is  a clear trend in the literature to develop models that have a huge number of parameters. This is making the computational cost and memory usage of ANNs prohibitive for smaller devices like tablet, smartphone and so on, not to mention the pollution aspect. In response to these issues, it is becoming necessary to train such models using numerical formats that use less bits i.e. that have a bigger machine error.
% The analysis of gradient type algorithms that takes into consideration this kind of errors is not well developed and present works often assume convexity of the objective function, an hypothesis that is commonly false for ANNs. We propose the use of a different hypothesis that is a well known generalization of convexity, called quasi-convexity.
% We develop both a deterministic and stochastic convergence analysis in this setting, furthermore we study the special case of ANNs giving specific bounds for these models.{\color{red} Maybe some emphasis on the fact that we provide theoretical justifications for convergence that it is already happening}
\end{abstract}

\section{Introduction}
\begin{figure} 
    	\centering
 \includegraphics[width=0.95\linewidth]{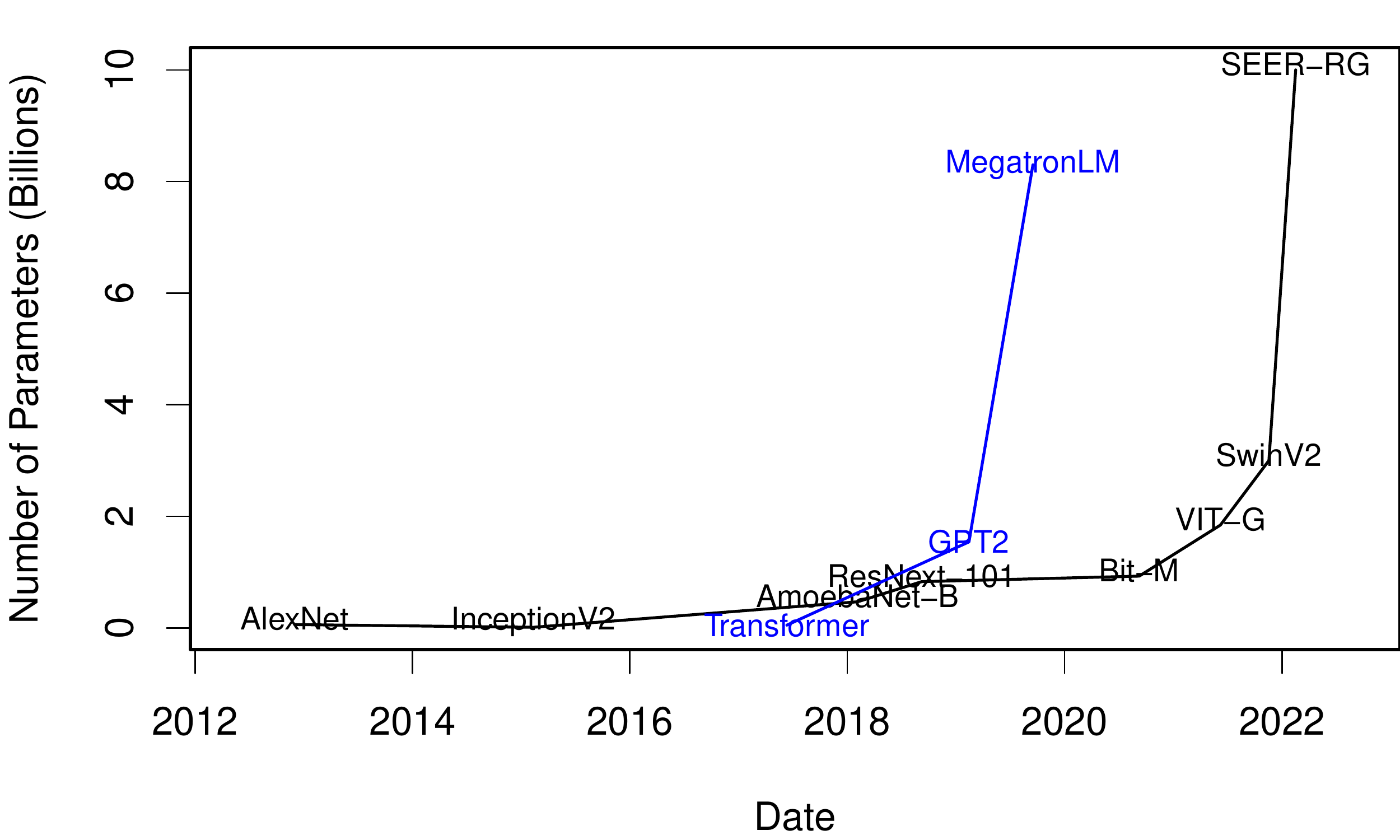}

    	\caption{Exploding trend of deep learning models for image classification (black) and language models  (blue) in time.}
    	\label{fig:trend}
    \end{figure}

\begin{figure} 
    	\centering
 \includegraphics[width=0.45\linewidth]{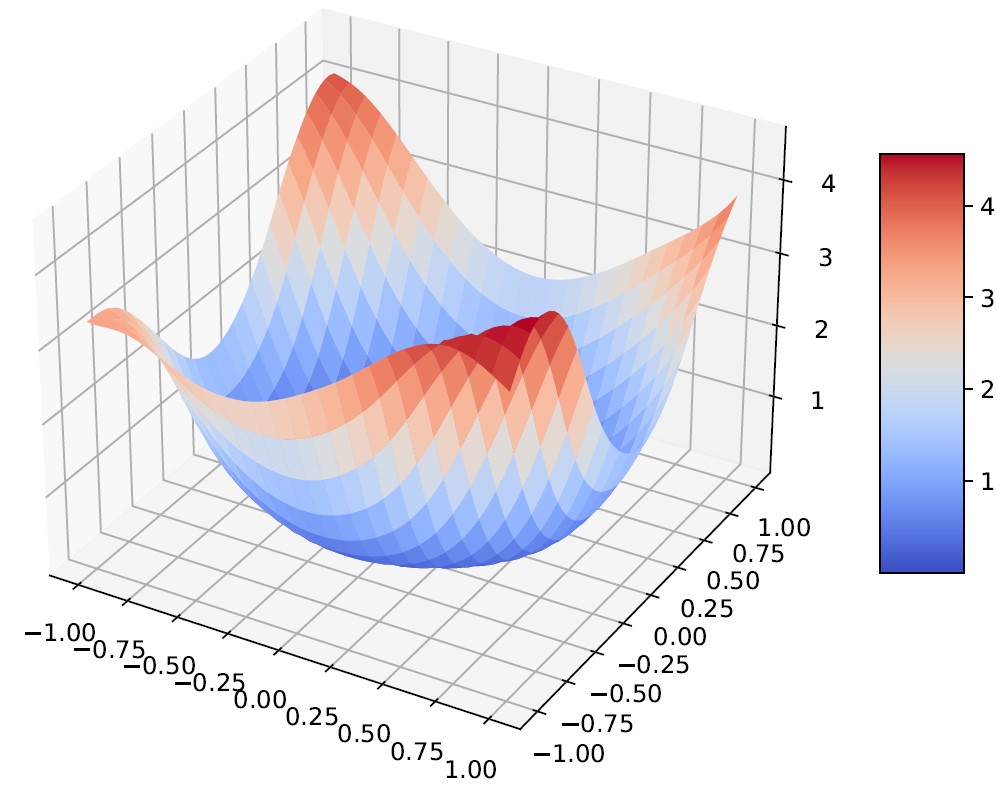}
\includegraphics[width=.45\linewidth]{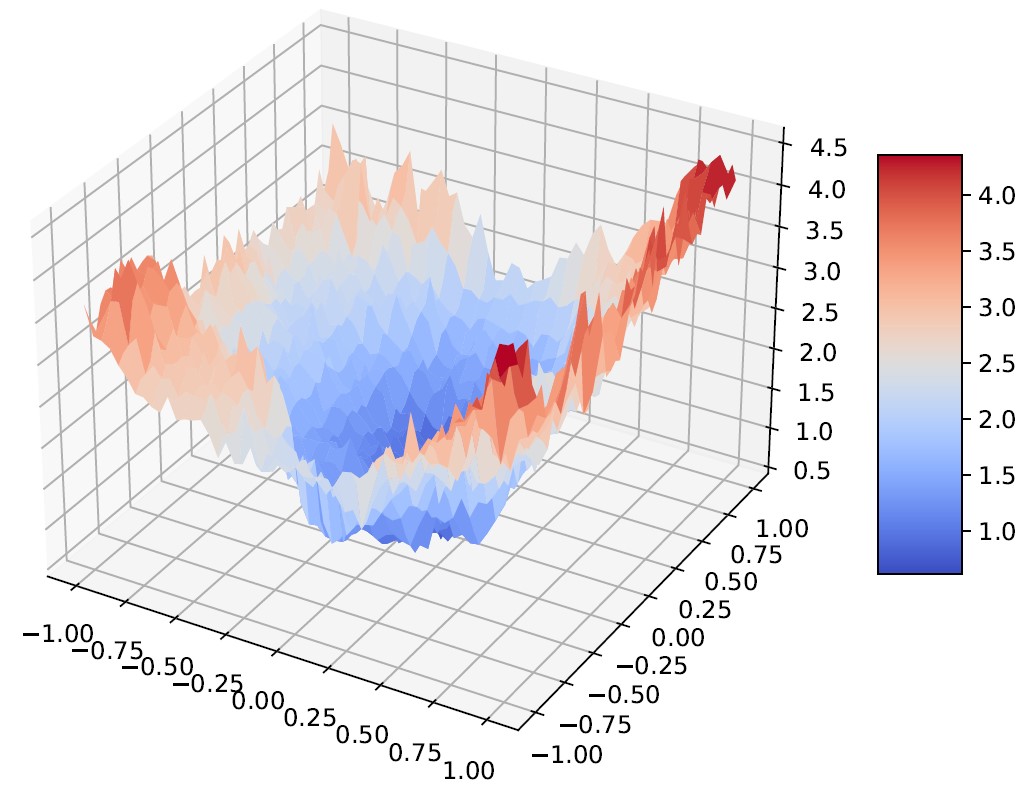}
    	\caption{ResNet-18 loss landscape in single-precision (left) and 
    	low-precision number format (right).}
    	\label{fig:low-bit-loss}
    \end{figure}

The success of deep learning models in different machine learning tasks have made these models de facto for almost all vision, text, and speech processing tasks. Figure \ref{fig:trend} depicts the size of deep learning models, indicating an exponential increase in the size of the models, and hence an urge for efficient computations. A common technique used in training deep learning models is SGD but the theoretical behaviour of SGD in rarely studied in low-precision number formats. Although there is a surge of articles on real numbers (for example see \cite{polyak1967}, \cite{schmidt2011}, \cite{ram2009}), the performance of SGD in low-precision number formats started recently. Depending on the precision, the loss landscape can change considerably. Figure \ref{fig:low-bit-loss}, for instance, depicts this situation for ResNet-18 loss landscape in both single-precision and low-precision number formats. Motivated by Figure \ref{fig:low-bit-loss}, we present a formal study of SGD for quasi-convex functions when computations are performed in low-precision number formats. 

We note that numerical errors, both in forward and back propagation, can possibly affect the convergence behaviour of the algorithm. It is conceivable that the numerical errors should increase as the precision decreases. To understand the effect of number format on the convergence of SGD, a careful analysis of the SGD algorithm for a predetermined precision is needed. We present both deterministic and stochastic analysis of the normalized SGD algorithm, obtaining bounds that show, explicitly, the effect of precision, i.e. number format. Such bounds can provide guidelines as to how SGD convergence is affected when constraints render performing high-precision computations impractical, and to what extent the precision can be reduced without compromising SGD convergence.  

Our experiments are performed for logistic regression on MNIST dataset. They confirm that the trajectory of the loss in low-precision SGD setup has at least a limit point whose loss value is in the proximity of the minimum when the numerical errors are relatively small, see Theorem~\ref{theorem1} and Theorem~\ref{theorem3}.

% based on the errors that we have for grads and accumulator, we can predict liminf of the loss trajectory.
% our results show that with stoch. bound we can predict closely the liminf, although with limitaion of estimating L and p, fp error delta (currently we have them in heuristic mode ....) 
% logistic reg on mnist, but mnist is reduced SGD, ca_linear
%  logistic reg on minist PCA, ...., this time we have error in grads

% computed the  optimum with GD, convex functions, 

% We start by recalling some standard result on quasi-convex functions. We then expand these results to analyze the convergence behaviour in the floating point environment and present some experimental results on different neural networks. 
%Finally we present some future directions.

This paper is organized as follows. Section \ref{sec:related-works} presents a literature review on the low-precision training of deep learning models and also provides some common background for theoretical analysis of SGD. Section \ref{sec:perlim} discusses some preliminary notations and definitions for analysis of quasi-convex loss function and also the floating point number formats. Section \ref{sec:main-res} contains the main theoretical results. 
%In Section \ref{sec:special-case} we study the bounds for a spacial case of convolutional neural networks. 
Section \ref{sec:experiments} provides some experimental results that support our theoretical results. We conclude in Section \ref{sec:conclusion}.
    
%\begin{figure} 
 %   	\centering
 %\includegraphics[width=0.95\linewidth]{imgs/arch.pdf}

  %  	\caption{Exploding trend of deep models for image classification (black) and language models  (blue) in time.}
   % 	\label{fig:trend}
    %\end{figure}

\section{Related works}\label{sec:related-works}
% \vahid{start with background on large models, low-bit hardware, and the trend. Argue for each definition if it is not standard SGD asumption and clarify the main result in the beggining.}.

Recently, deep learning models provide state-of-the-art performance in various machine learning tasks such as computer vision, speech, and natural language processing (NLP). The size of ImageNet classification models after the introduction of AlexNet size is  exploded to $200\times$, and the size of language models are  getting $10\times$ bigger every year. The recent trend of deep learning models shows that larger models such as transformers \citep{vaswani2017attention} and their variants such as GPT2 \citep{radford2019language}, MegatronLM \citep{shoeybi2019megatron}, and  \cite{brown2020language}  are easier to generalize on different downstream tasks. Moreover, examples of large language models are included in Figure~\ref{fig:trend} (blue line) and they show an increasing trend in number of parameters over time. A similar trend also appear in vision models, specially after the advent of vision transformers \citep{zhai2022scaling, goyal2022vision} that beat convolutional neural networks \citep{mahajan2018exploring} on various tasks, see Figure~\ref{fig:trend} (black line). Although such large models have advantage in terms of accuracy, they suffer from high computational cost in their training and inference phases. Moreover, the high computational complexity of these models causes high energy consumption and memory usage which makes their training and deployment difficult and even sometimes infeasible. Thus, reducing the computational complexity of large deep learning models is crucial.

On the other hand, there has been some efforts in manually redesigning smaller models with similar accuracy as large models which often require more complicated training. In image classification small models such as MobileNet \citep{howard2017mobilenets} have a similar accuracy as ResNet \cite{he2016deep}, and in language models, DistilBERT \citep{sanh2019distilbert} shows close performance to BERT. Meanwhile there have been some efforts in designing models automatically  such as \cite{liu2018darts, zoph2018learning}. Other methods include those preserving the baseline model's architecture while modifying computations e.g.  compressing large models using sparse estimation \cite{luo2017thinet, ramakrishnan2020differentiable,furuya2022spectral}, or simplifying computations by running on low-precision number formats \cite{jacob2018quantization, wu2020integer}. Some researchers are even pushing frontiers by storing weights and reducing activation to binary \citep{hubara2016binarized} or ternary numbers \citep{li2021s}.

Training large models are compute intensive using single-precision floating point. This is why hardware manufacturers such as NVIDIA, Google, Transcent, and Huawei started supporting hardware for low-precision number formats such as Bfloat, float16, and int8.  Recently researchers try to map single-precision computations on lower bits, see \cite{zhang2020fixed, zhao2021distribution, ghaffari2022integer}.

% Deep learning models are reaching to a point that their resource requirement are becoming overwhelming such that only big companies can afford them.  

Majority of the literature on SGD assumes convex loss function. We weaken this assumption by considering quasi-convex class of loss functions that include convex functions as special case. One of the first works on quasi-convex optimization is \citet{kiwiel1996}, where is proven that the gradient descent algorithm converges to a stationary point.
Later, in \citet{kiwiel2001}, the differentiability hypothesis is removed and the convergence result is shown using quasi-subgradients. In the case of perturbed SGD, in \citet{hu2015} the authors are able to deal with bounded biased perturbation on the quasi-subgradient computation. In a subsequent work  \citet{hu2016} analyzed the stochastic setting. Recently, \cite{zhang2022low} studied the low-precision SGD for strongly convex loss functions where the authors used Langevin dynamics. In comparison, our work differs in two aspects (i) we assume quasi-convexity, (ii) our setup adds noise to the SGD and this allows for less stringent assumptions on the noise and its distribution.
%{\color{red}Need more literature for: floating points GD, papers that show training in low bit works}

\section{Preliminaries}\label{sec:perlim}
We start with some preliminary notations about quasi-convexity and floating point number formats in the sequel.

\subsection{Quasi-convexity}

\begin{definition}
A function $f:\RRn\rightarrow \RR$ is said to be quasi-convex if $\forall a\in \RR$, $f^{-1}[(-\infty,a)]=\{\x \in \RRn|\,f(\x)\in(-\infty,a)\}=\S_{f,a}$ is convex.
\end{definition}

% \begin{figure}
%     	\centering
%     	\includegraphics[width=3cm]{QC_notQC.PNG}
%     	\caption{A quasi-convex function that is not convex (on the left) and a function that is not quasi-convex (on the right)}
%     \end{figure}

\begin{definition} %[quasi-subgradient, \cite{hu2015}]
Given a quasi-convex function $f:\RRn\rightarrow \RR$, the quasi-subgradient of $f$ at $\x\in\RRn$ is defined as $\bar{\partial}^* f(\x)=\{g\in \RRn \mid \langle g,\x'-\x\rangle\leq0,\; \forall \x' \in \S_{f,f(\x)}\} $
\end{definition}

In what follows, the optimal value and optimal set of a function $f$ on a set $\X$ are respectively denoted by $f^*$ and $\Xs$, i.e. $f^*=\inf_{\x \in \X} f(\x)$ 
%the optimal value  of $f$ and with 
and $\Xs=\argmin_{\x \in \X} f(\x) .$ 
%the optimal set of the minimization problem over $f$.

\begin{definition}
Let $p>0$ and $L>0$. $f:\RRn\rightarrow \RR$ is said to satisfy the H$\ddot{\text{o}}$lder condition of order $p$ with constant $L$ if
\[
f(\x)-f^*\leq L[\dist(\x,\Xs)]^p
\]
where $\dist(\x,\x')=\min_{\x'} \|\x-\x'\|$ where $\|\cdot\|$ denotes the Euclidean norm.
\end{definition}

\subsection{Floating points}
\label{floats}
%We will now present the basic notions of a floating point environment.

A base $\beta\in\NN$, with precision $t\in\NN$, and exponent range $[e_{\min},e_{\max}]\subset\ZZ$ define a floating point system $\FF$, where an element $y\in\FF$ can be represented as  
\begin{alignat}{6}\label{fp}
y=\pm m\times \beta^{e-t},
\end{alignat}
where $m\in\ZZ$, $0\leq m<\beta^t$, and $e\in [e_{\min},e_{\max}]$.

For an $x\in [\beta^{e_{\min}-t},\beta^{e_{\max}}(1-\beta^{-t})]$, let the float projection function be  $\fl(x):=\mathsf{P}_\FF(x)$, then for $x,y\in\FF$, the rounding error for basic operations $\op\in\{+,-,\times,/\}$, is 
\begin{alignat}{6}
\fl(x\op y)=(x\op y)(1+\delta_0), \label{operror}
\end{alignat}
where the error is bounded by $\delta_0\leq \beta^{1-t}$ . 

When trying to compute a subgradient $g\in\partial f(w)$ for a $w\in\FF$ the error is bounded to 
\begin{alignat}{6}
\fl(g)=g(1+\delta_1),\nonumber
\end{alignat}
where $|\delta_1|\leq c \delta_0$ and $c>0$ depends on the number of operations needed for computing such subgradient.
A step of subgradient descent in floating point in $\RRn$ is:
\begin{alignat}{6}
&\w_{k+1}&&=\left\{\w_{k}-\t [\g_k(1+\boldsymbol\delta_{1k})]\right\}(1+\boldsymbol\delta_{2k}),\nonumber
\end{alignat}
where we suppose $\eta \in \FF,$ $\norm{\boldsymbol \delta_{1k}}\leq c\delta_0\sqrt{\n}$ and $\norm{\boldsymbol\delta_{2k}}\leq \delta_0\sqrt{\n}$ where $\n$ is the dimension of $\boldsymbol \delta$. 
We can reformulate it in terms of absolute error
\begin{alignat}{6}
&\x_{k+1}&&=\x_{k}-\eta(\g_k+\r_k)+\s_k,\nonumber
\end{alignat}
and if  the norm of the subgradient $\g_k$ and $\w_{k+1}$ is bounded, then so are $\norm{\r_k}\leq R$ and $\norm{\s_k}\leq S$.
Note that the infinity norm of the errors are bounded
\[
\|\delta_{1k}\|_\infty\leq c\delta_0, \qquad \|\delta_{2k}\|_\infty \leq \delta_0,
\]
so
\[
{R}^2\leq c\delta_0 \norm{\g_k}, \quad
S^2 \leq \delta_0\norm{\x_k}.
\]
%This is important because the final term does not depend on the dimension $\n$.

\section{Main results}\label{sec:main-res}
Although our study is mainly motivated by training deep learning models and floating point errors, they can be applied elsewhere. 
%can be applied in any context where the objective function is quasi-convex and there is an error both in the gradient computation and on the summation of it with the last point.

\subsection{Deterministic analysis}
%We start with a worst case scenario analysis.
%Form now on we denote with $f^*=\inf_\x f(\x)$ the optimal value  of $f$, with $\Xs=\argmin_\x f(\x)$ the optimal set of the minimization problem over $f$.
%In some cases we will also consider to be known a set $\Xs$ such that $\Xs\subset \X$ and we denote with  $\piX(\cdot)$ the projection operator over $\X$.

%The first lack in the literature is the analysis of the case when there is an error in the sum among the gradient and the last iterate computation. 
Let $\piX(\cdot)$ be the projection operator over $\X$. We start with adapting and improving the results of \citet{hu2015} in the presence of error in the summation

\begin{theorem}\label{theorem1}
Let $f:\RRn\rightarrow \RR$ be a quasi-convex function satisfying the H$\ddot{\text{o}}$lder condition of order $p$ and constant $L$. Let $\x_{k+1}=\piX(\x_k-\t\hat{g}_k+\s_k)$ where $\X\subset \RRn$ is compact, $c$ is the diameter of $\Omega$, and $\hat{\g}_k=\frac{\g_k}{\|\g_k\|}+\r_k$, with $\g_k \in \bar{\partial}^*f(\x_k)$,  $\|\r_k\|<R<1, \; \|\s_k\|\leq S $. Then 
\[
\liminf_{k\rightarrow \infty} f(\x_k)\leq f^*+L\Gamma^p(c), 
\] 
where

{\footnotesize
$$
\hspace{-.1 in}\Gamma(c)=\frac{\t}{2}\left[1+\left(R+\frac{S}{\t}\right)^2\right] \bigvee \left[ \frac{\t}{2}\left\{1-\left(R+\frac{S}{\t}\right)^2\right\}+c \left( R +\frac{S}{\t}\right)\right]. 
$$}
\end{theorem}
See the Appendix for the proof.\newline
{\bf Remark:} Unlike \citet{hu2015}, decreasing $\eta$ does not decrease  the error bound always, so we can derive its optimal value by minimizing the bound with respect to $\t$. 

Define 
\[
\t_1=\frac{S}{\sqrt{1+R^2}}, \quad  
\t_2=\sqrt{\frac{S(c-2S)}{1-R^2}}, \quad \t_3=\frac{c-S}{R}
\]
\begin{corollary}\label{corollary1}
The optimal choice for the step size $\t$ that minimizes the error bound in Theorem \ref{theorem1} is reached in at least one of this 3 points $\{\t_1,\t_2,\t_3\}$.
\end{corollary}

%The results we just proved are can be used in floating point scenario, since $\norm{\hat\g_k}=1$ we can compute $R$. Also, thanks to the projection, we can bound the norm of the next iterate and compute $S$.

%If we run the algorithm for the total of $K$ iterations, since the result is about the $\liminf$ of the trajectory, using the last $\x_K$ as final solution could not be the best idea. We want instead to keep track of the $\x_k$ with the best objective value, that is $\x_j$ with  $j=\argmin\{f(\x_k)|\, k\leq K\}$.
%We can adapt the previous result \vahid{which result?} to obtain an upper bound on the number of iterates needed to visit at least a point with a certain objective value.

% \begin{theorem} \label{theorem2}
% Let $f:\RR^n\rightarrow \RR$ be a quasi-convex function satisfying the Holder's condition of order $p$ and constant $L$. Let $y_{k+1}=x_k-t\hat{g}_k$ and 
% $x_{k+1}=y_{k+1}+s_k$ where $\hat{g}_k=\frac{g_k}{\|g_k\|}+r_k$, with $g_k \in \bar{\partial}^*f(x_k)$, with $\|r_k\|<R, \; \|s_k\|\leq S \;\forall k$. Then, $\forall\; \delta>0$,
% \begin{equation*}
%     \min_{k<\kappa(\delta)} f(x_k)\leq f^*+L(\Gamma(D_0)+\delta)^p
% \end{equation*}
% with $\kappa(\delta)=\frac{D_0}{2t\delta}$, $D_0=\|x_0-x^*\|$  
% \end{theorem}

The next result presents a finite iteration version of the previous result. The effect of the number of iterations $K$, and the starting point $\x_0$ are clearly reflected in the bound for $\min_{k<K} f(\x_k)$.

\begin{theorem} \label{theorem2}
Let $f:\RRn\rightarrow \RR$ be a quasi-convex function satisfying the H$\ddot{\text{o}}$lder condition of order $p$ and constant $L$. Let $\x'_{k+1}=\x_k-\t\hat{\g}_k$ and 
$\x_{k+1}=\x'_{k+1}+\s_k$ where $\hat{\g}_k=\frac{\g_k}{\norm{\g_k}}+\r_k$, where $\g_k \in \bar{\partial}^*f(\x_k)$, $\|\r_k\|<R, \; \|\s_k\|\leq S.$ Then,
\begin{equation*}
    \min_{k<K} f(\x_k) \leq f^*+L\left[\Gamma(c_0)+\frac{c_0}{2\eta K}\right]^p,
\end{equation*}
with $c_0=\|\x_0-\x^*\|.$  
\end{theorem}
See the Appendix for the proof.\newline

\subsection{Stochastic analysis}

% In many contexts, in particular in deep learning training, the complete gradient information is not available. In such scenarios we may only have access to a stochastic gradient. 
Here, we present the stochastic counterpart of Theorem \ref{theorem1}. The theorem requires only mild conditions on the first two moments of the errors. We start by defining the notion of stochastic quasi-subgradient. 
%For the quasi-convex case the following definition is useful to generalize the concept of stochastic gradient.

\begin{definition} \label{definition1}
Let $\x$ and $\x'$ be $d$-dimensional random vectors defined on the probability space $(\mathcal{W}, \F, \mathcal{P})$ and $f:\RRn\rightarrow \RR$ be a measurable quasi-convex function. Then $\g(\x)$ is called a unit noisy quasi-subgradient of $f$ at $\x$ if 
%$\EE\left\{\g(\x) \}\in \bar{\partial}^* f(\x),$ that is 
$\| \g(\x) \| \stackrel{a.s.}{=}1$ and 
% \mathcal{P}  \left \{w \in \S_{f,f(w)}: \EE\left\{\langle  g(\x), \x'-\x\rangle | \x' = w \}> 0 \right\} = 0.
$\mathcal{P} \left \{\S_{f,f(\x)} \cap \mathcal{A}_{\x} \ne \emptyset \right \} = 0, $
where $\mathcal{A}_{\x} = \{ \x' : \langle  \g(\x), \x' - \x \rangle >  0 \} \cdot $
\end{definition}
Thus, inspired by results of \citet{hu2016}, we prove the following theorem that take into account both randomness in the gradient and the computation numerical error.

\begin{theorem}\label{theorem3}
Let $f :\RRn\rightarrow \RR$ be a continuous quasi-convex function satisfying the H$\ddot{\text{o}}$lder condition of order $p$ and constant $L$ . Let $\x_{k+1}=\piX[\x_k-\t(\hat{\g}_k+\r_k)+\s_k]$ where $\X$ is a convex closed set, $\hat{\g}_k$ is a unit noisy quasi-subgradient of $f$ at $\x_k$, $\r_k$'s are i.i.d. random vectors with $\EE\left\{\r_k\right\}=\mathbf 0$ and $\EE\left\{\|\r_k\|^2\right\}=\V$, and similarly $\s_k$'s are i.i.d random vectors  with  $\EE\left\{\s_k\right\}=\mathbf 0$ and $\EE\left\{\|\s_k\|^2\right\}=\W$. Further assume that $\hat{\g}_k$, $\s_k$, and $\r_k$ are uncorrelated. Then, 
$$\liminf_{k\rightarrow \infty} f(\x_k)\leq f^*+L\left[\frac{\t}{2}(1+\V)+\frac{\W}{2\t}\right]^p
\quad a.s.$$
\end{theorem}
See the Appendix for the proof.\newline

\begin{figure}
\centering
        \includegraphics[width=0.7\linewidth]{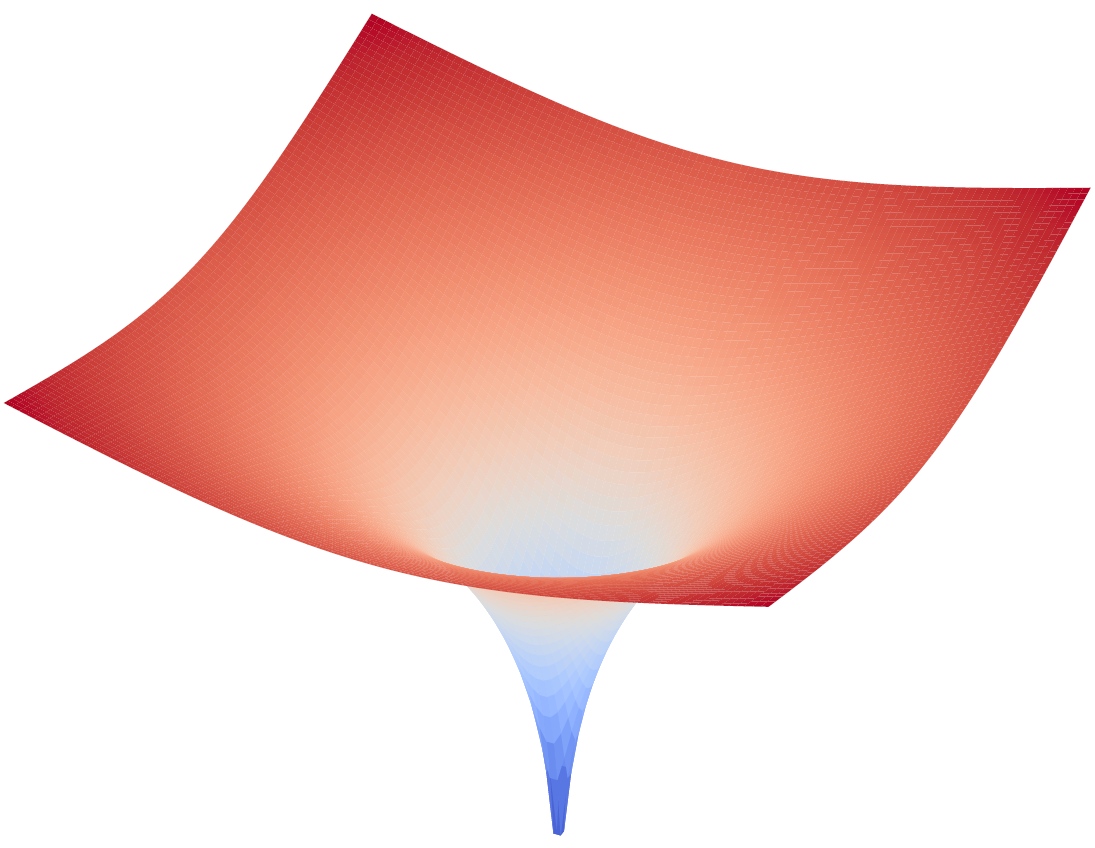}\\
        \includegraphics[width=\linewidth]{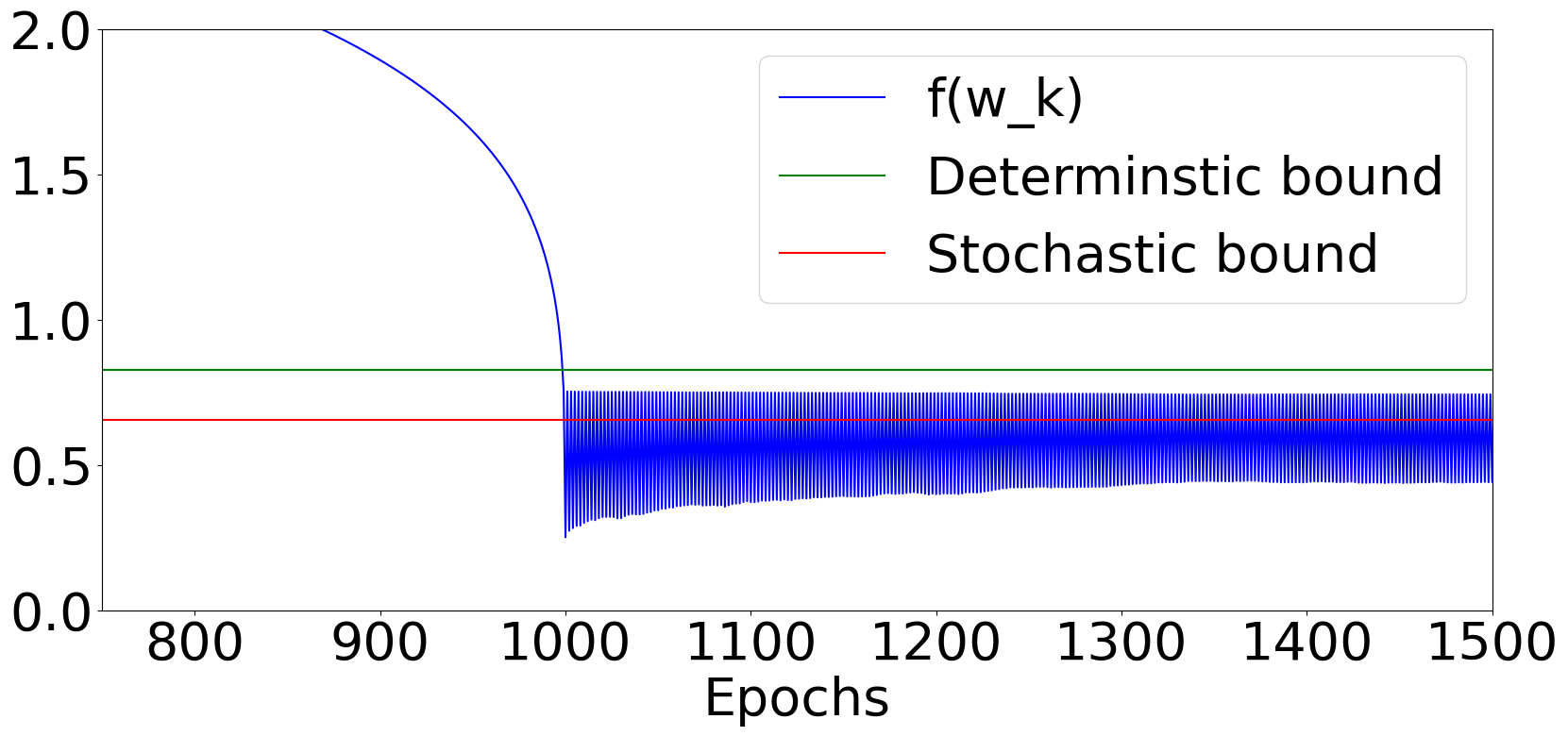}
    	\caption{A two dimensional $\norm{\w}^{0.2}$ quasi-convex function (top panel), and  the SGD trace plot confirming the stochastic and deterministic bounds hold (bottom panel).}
    	\label{fig3}
    \end{figure}

Similar to Corollary \ref{corollary1} one can derive the optimal step size.
\begin{corollary}
\label{corollary2}
The optimal step size $\t$ that minimizes the error bound in Theorem~\ref{theorem3} is  $\t^*=\sqrt{\frac{\W}{\V+1}}.$
\end{corollary}

\textbf{Remark:} It immediately follows the optimal step size is $\frac{\sigma_s}{\sigma_r}$ for large $d$.

\section{Experiments}\label{sec:experiments}

We performed two types of experiments, a simple quasi-convex function and a logistic regression on MNIST dataset.

\subsection{Simple quasi-convex function}
 To asses the bounds obtained in our theorems we start with a simple quasi-convex function that exactly satisfy the Holder's condition. We chose $f(\x)= 3 \|\x\|^{0.2}$ where $\x\in \RR^{40}$. In this example, the parameters of Holder's condition are $p=0.2$ and $L=3$. We added noise to the gradients  and to the weight update denoted by $\| \r_k \|$ and $\| \s_k \|$ respectively. This noise has a uniform distribution $\r_{ki}\sim U(-B_r, B_r),$ and $s_{ki}\sim U(-B_s, B_s)$. Figure \ref{fig3} shows the stochastic and deterministic bound for this experiment. Note that, the theoretical bounds holds in both stochastic and deterministic cases.
%  \begin{itemize}
%     \item The bounds are respected
%     \item Stochastic bound is on point, worst case obviously not.
%     \item Our predicted optimal step size is actually working
%     \item Something on convergence speed?
% \end{itemize}

\subsection{Logistic regression}
\begin{figure}
        \includegraphics[width= 0.45\linewidth]{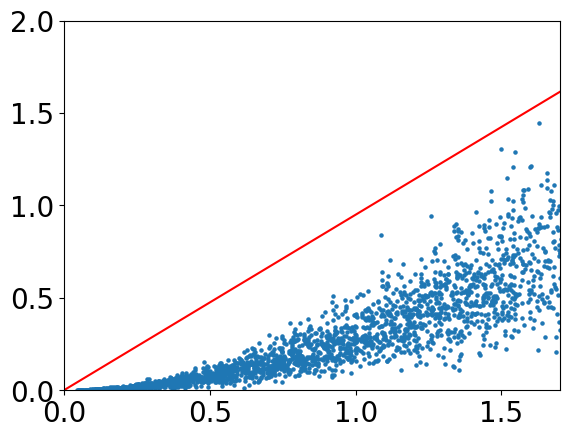}
         \includegraphics[width=0.45\linewidth]{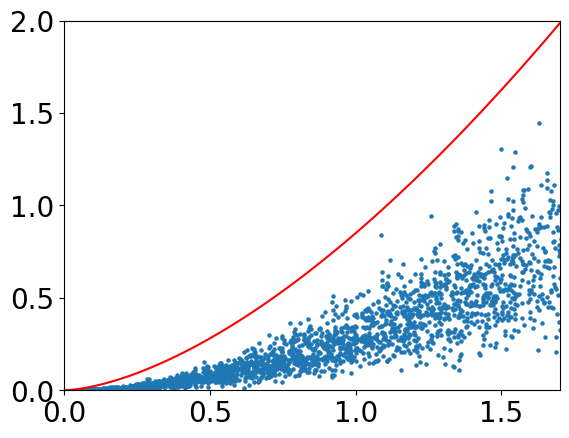}
    \caption{ The Holder's parameters $p$ and $L$ are manually fitted to the loss function that is evaluated at different distances from the optimal point. The left panel is a linear fit with $p=1$ and $L=0.95$. The right panel demonstrate the fit with $p=1.6$ and $L=0.85$. We used $p=1.6$ and $L=0.85$ for our experiments.}
    	\label{fig:l-p-fit}
    \end{figure}

Here, we present experimental results of logistic regression on the first two principal components of MNIST dataset. For this experiment, we need to estimate the parameters of the Holder's condition for the loss function in order to compute the bounds. To do so, the Holder's parameters $p$ and $L$ are manually fitted to the loss function that is evaluated at different distances from the optimal point, see Figure \ref{fig:l-p-fit}. 
The optimal point $\w^*$ in our experiments is obtained using single-precision floating point gradient descent (GD) method and is used as a reference to compute the parameters of the bounds $f^*$ and $c$. \
% Since the loss function of a logistic regression is convex, its optimal point is unique and the algorithm does not converge to a local optimum point. 

Computation of the gradients involves inner products that are computed by multipliers and accumulators. The accumulator have numerical error relative to its mantissa size. We tested our logistic regression setup using Bfloat number format and reduced accumulator size. Also note that according to Theorem \ref{theorem3}, the values of $\W$ and $\V$ are required to compute the bounds. Thus, in our experiments, we used empirical values of those parameters to compute the bounds. Also note that we did not plot the deterministic bounds for these experiments as they are too pessimistic.

    \begin{figure}
        \includegraphics[width=0.95\linewidth]{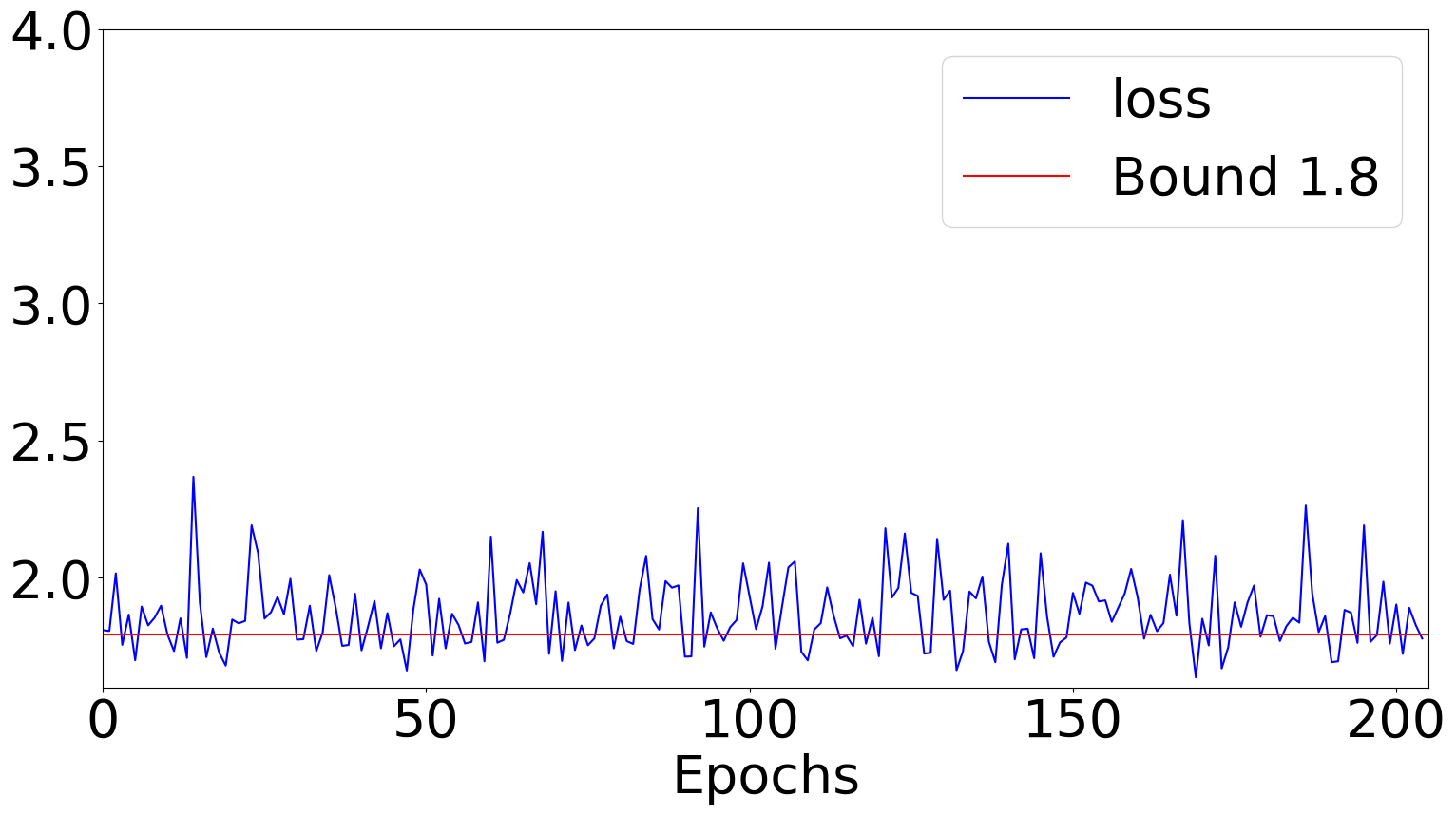}
    	\caption{Logistic regression trained using single-precision SGD and a fixed learning rate.}
    	\label{fig:noErr}
    \end{figure}

\begin{figure}
        \includegraphics[width=0.95\linewidth]{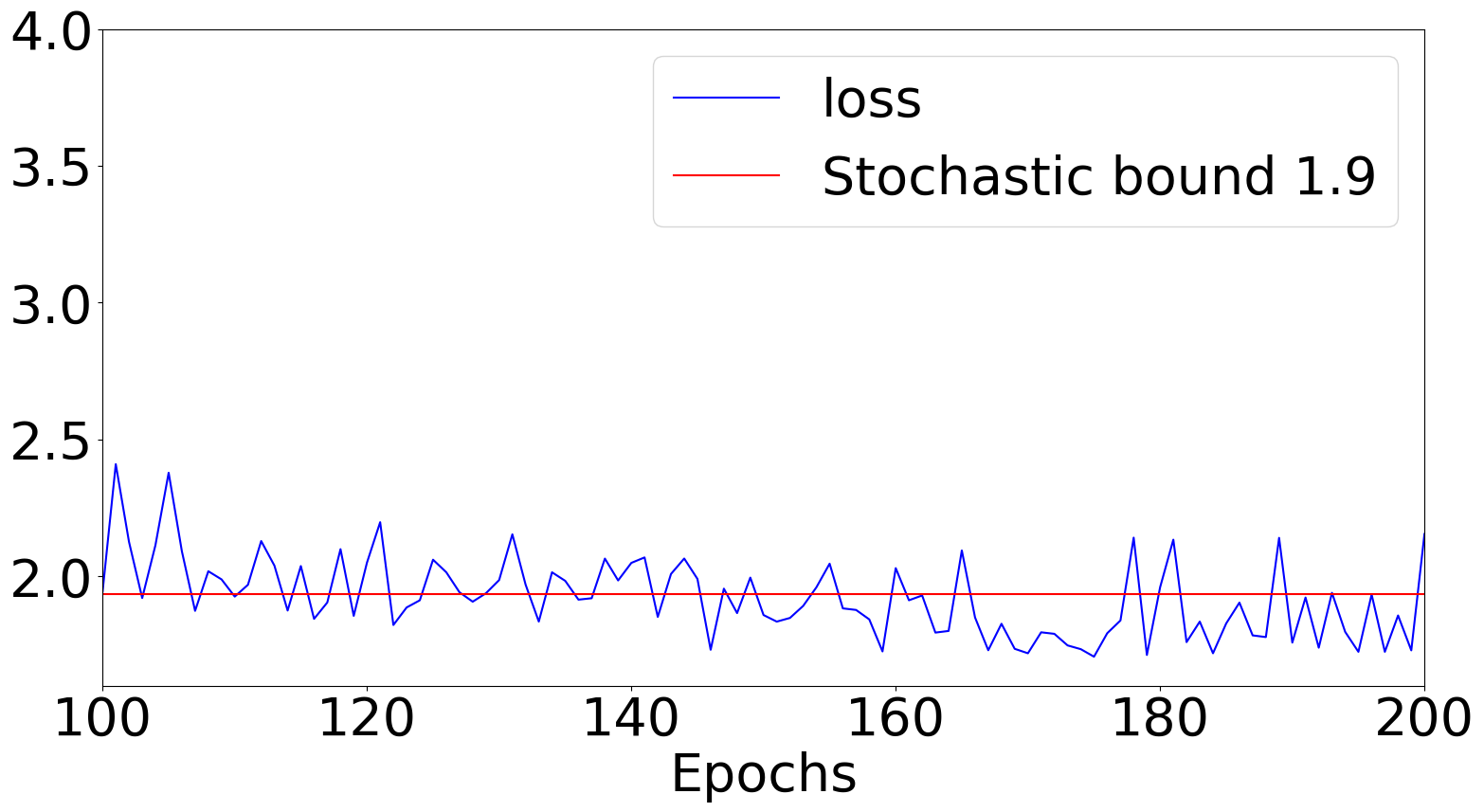}
    	\caption{Logistic regression trained using Bfloat SGD with accumulator size of 15 ($\s_k \neq 0$ and $\r_k\neq 0$).}
    	\label{fig6}
    \end{figure}

\begin{figure}
        \includegraphics[width=0.95\linewidth]{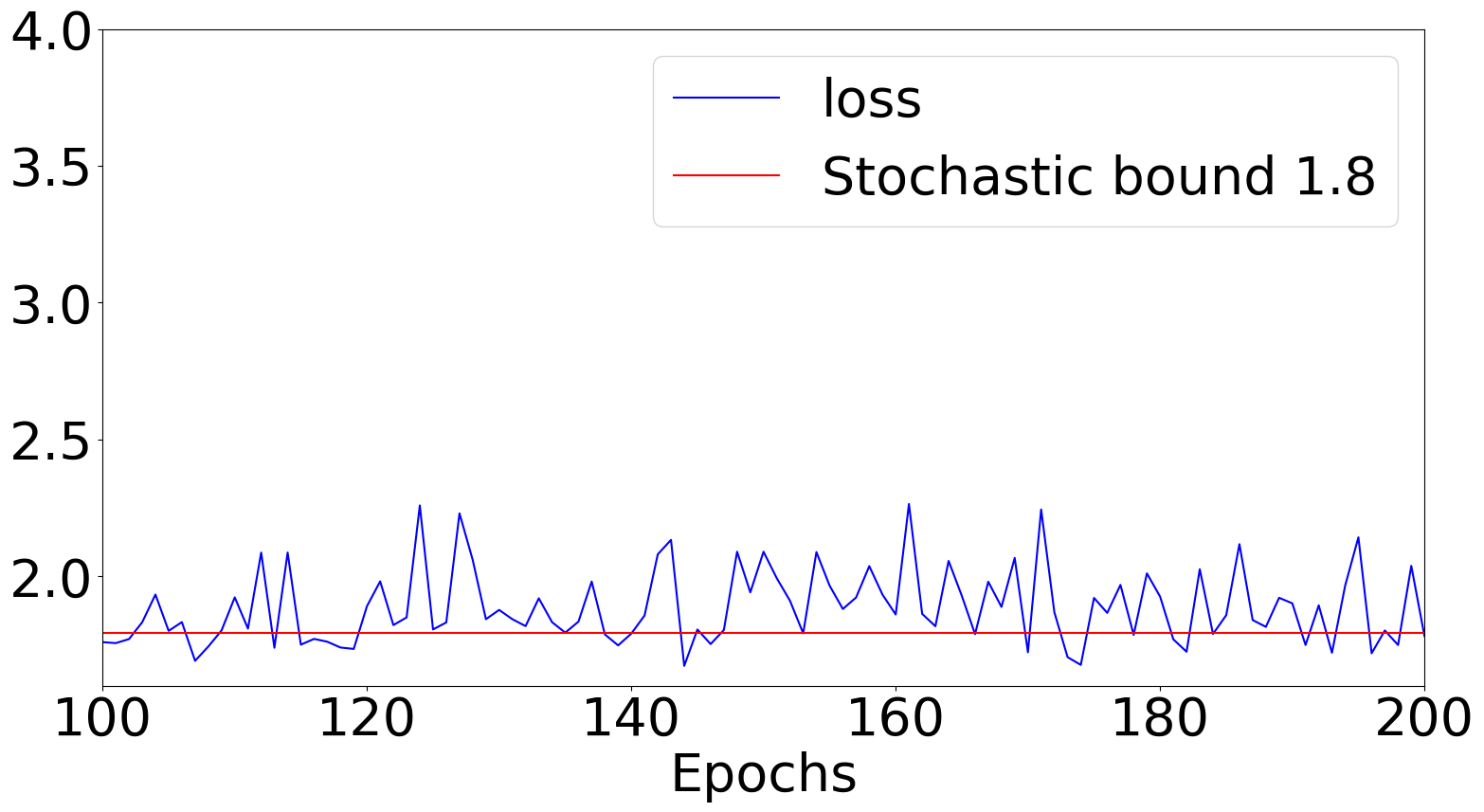}
    	\caption{Logistic regression trained using Bfloat gradients with accumulator size of 15, and single-precision weight update ($\s_k = 0$ and $\r_k\neq 0$).}
    	\label{fig7}
    \end{figure}

In order to evaluate the Holder's condition parameters, $p$ and $L$, estimated as shown in Figure \ref{fig:l-p-fit}, we use a single-precision SGD to confirm if the bounds hold. Figure \ref{fig:noErr} demonstrates that the loss trajectory (blue line) has a limit point in the proximity of the optimal point of the convex loss function.
Figure \ref{fig6} demonstrates the loss trajectory when both weight update and gradient computations are performed using Bfloat number format. Note that Bfloat has 8 bit exponent and 7 bit mantissa and is used recently to train deep learning models. The computations are performed using 15-bit accumulator mantissa. Figure \ref{fig7} shows the loss trajectory when the weight update is in single precision and only gradient computations are performed using Bfloat number format. In this experiment, the stochastic bound is numerically equal to the single-precision SGD, indicating that the precision of weight update is more important compared to the precision of the gradients. 

    \begin{figure}[!t]
        \includegraphics[width=0.95\linewidth]{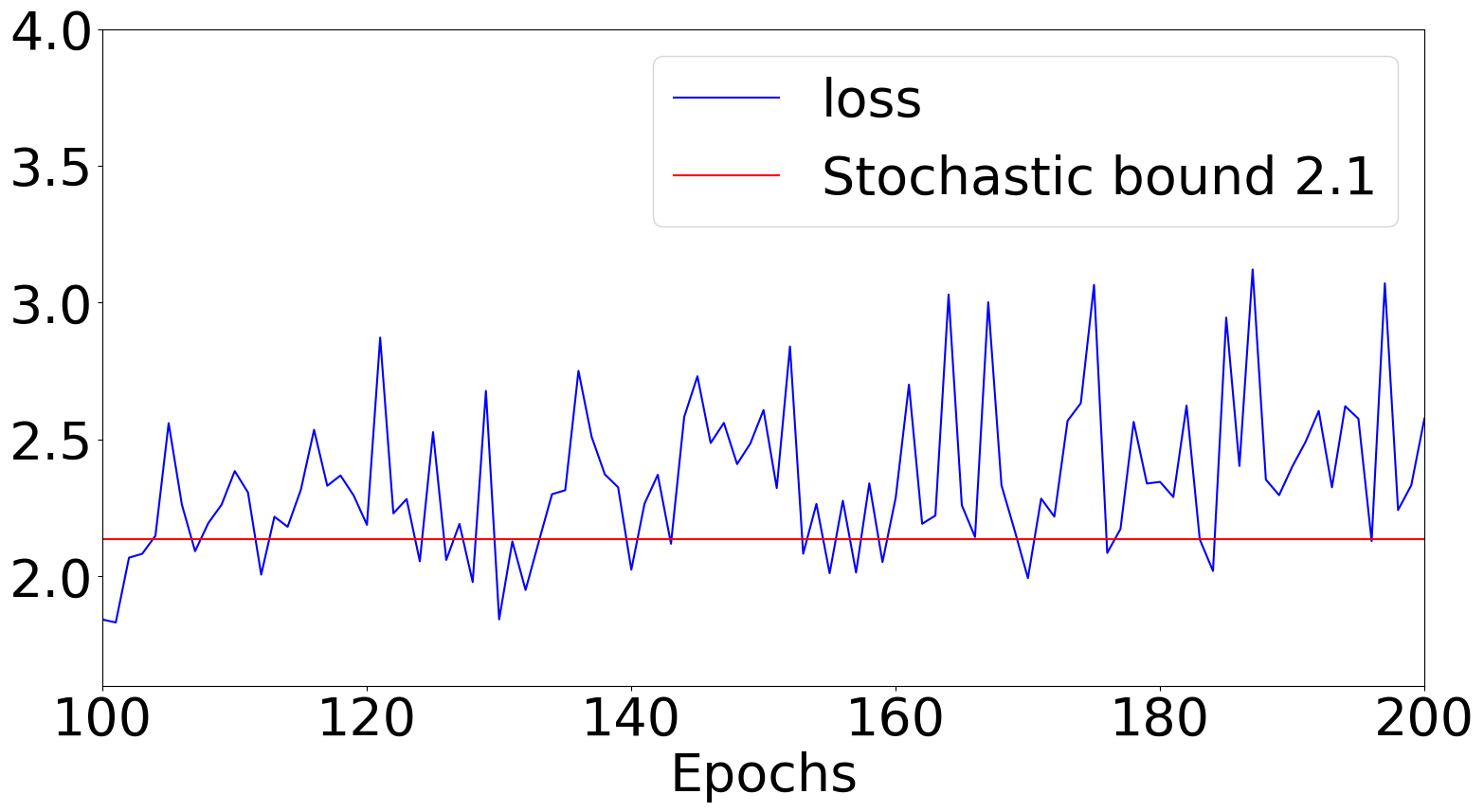}
    	\caption{Logistic regression trained using Bfloat gradients with accumulator size of 10, and single-precision weight update ($\s_k = 0$ and $\r_k\neq 0$).}
    	\label{fig8}
    \end{figure}  
    
Reducing the accumulator mantissa size has a direct effect on the convergence of SGD. Figure \ref{fig8} shows that stochastic bound is increased in the case of 10-bit accumulator size. In this experiment, the loss trajectory oscillates more in the neighbourhood of the optimum point. This indicates the accumulator size plays an important role in reducing the numerical errors of the low-precision SGD computations, and consequently improves the convergence of SGD.

% different learning rates and number of bits for the mantissa and accumulator. No rounding at accumulator is performed.

% The choice of $d$ for the worst case bound is crucial and heavily depends on our knowledge on the problem, we used $d=3\sqrt{n}$.

% Bounding the floating point error was very complicated, depending on the region the current point is located the error can vary a lot, also the error components are not completely unbiased. We provide some methods to estimate these errors but they can not be tight for the mentioned reasons, so when reporting our results we used the empirical error measured during the experiments (a 32 bit floating point model was used as a reference).

% Since we estimate $p$ and $L$ we first provide some results in a full precision setting to show that the (already well known in the literature) result holds in the setting without perturbations \ref{fig:noErr}.

% \begin{itemize}
%     \item Experimental setting: Logistic regression on reduced MNIST dataset.
%     \item Values for $p$, $L$, $f^*$, $d$, $R$, $S$, $W$, and $V$.
%     \item low bit training using Huawei framework
%     \item performance evaluation?
%     \item comparison with sgd or with other bounds?
% \end{itemize}

    % \begin{figure}
    %     \includegraphics[width=0.95\linewidth]{CNTRL_lreg2_lr1.0_e250_chkp0_d1_capnsgd_bit8_1665594542.png}
    % 	\caption{Logistic regression trained using Bfloat SGD with accumulator size of 10 ($s_k \neq 0$ and $r_k\neq 0$).}
    % \end{figure}  

\vspace{-.1 in}
\section{Conclusion}\label{sec:conclusion}

We have studied the convergence of low-precision floating-point SGD for quasi-convex loss functions and extended some existing deterministic and stochastic bounds for convex loss functions. In our theoretical setup, we considered numerical errors for weight update and gradient computations. We have also derived the optimal step size as a corollary of our theoretical results. Furthermore, in our experiments, the effect of numerical errors on weight update and gradient computations are demonstrated. Our experiments show that the accumulator mantissa size plays a key role in reducing the numerical error and improving the convergence of SGD. Although our experiments with logistic regression are promising, extension of the experiments for more complex models is an appealing direction as the future work.

%\subsubsection*{Acknowledgements}

\bibliography{aistat}

\begin{thebibliography}{30}
\providecommand{\natexlab}[1]{#1}
\providecommand{\url}[1]{\texttt{#1}}
\expandafter\ifx\csname urlstyle\endcsname\relax
  \providecommand{\doi}[1]{doi: #1}\else
  \providecommand{\doi}{doi: \begingroup \urlstyle{rm}\Url}\fi

\bibitem[Brown et~al.(2020)Brown, Mann, Ryder, Subbiah, Kaplan, Dhariwal,
  Neelakantan, Shyam, Sastry, Askell, et~al.]{brown2020language}
Tom Brown, Benjamin Mann, Nick Ryder, Melanie Subbiah, Jared~D Kaplan, Prafulla
  Dhariwal, Arvind Neelakantan, Pranav Shyam, Girish Sastry, Amanda Askell,
  et~al.
\newblock Language models are few-shot learners.
\newblock \emph{Advances in neural information processing systems},
  33:\penalty0 1877--1901, 2020.

\bibitem[Furuya et~al.(2022)Furuya, Suetake, Taniguchi, Kusumoto, Saiin, and
  Daimon]{furuya2022spectral}
Takashi Furuya, Kazuma Suetake, Koichi Taniguchi, Hiroyuki Kusumoto, Ryuji
  Saiin, and Tomohiro Daimon.
\newblock Spectral pruning for recurrent neural networks.
\newblock In \emph{International Conference on Artificial Intelligence and
  Statistics}, pages 3458--3482. PMLR, 2022.

\bibitem[Ghaffari et~al.(2022)Ghaffari, Tahaei, Tayaranian, Asgharian, and
  Nia]{ghaffari2022integer}
Alireza Ghaffari, Marzieh~S Tahaei, Mohammadreza Tayaranian, Masoud Asgharian,
  and Vahid~Partovi Nia.
\newblock Is integer arithmetic enough for deep learning training?
\newblock \emph{arXiv preprint arXiv:2207.08822}, 2022.

\bibitem[Goyal et~al.(2022)Goyal, Duval, Seessel, Caron, Singh, Misra, Sagun,
  Joulin, and Bojanowski]{goyal2022vision}
Priya Goyal, Quentin Duval, Isaac Seessel, Mathilde Caron, Mannat Singh, Ishan
  Misra, Levent Sagun, Armand Joulin, and Piotr Bojanowski.
\newblock Vision models are more robust and fair when pretrained on uncurated
  images without supervision.
\newblock \emph{arXiv preprint arXiv:2202.08360}, 2022.

\bibitem[He et~al.(2016)He, Zhang, Ren, and Sun]{he2016deep}
Kaiming He, Xiangyu Zhang, Shaoqing Ren, and Jian Sun.
\newblock Deep residual learning for image recognition.
\newblock In \emph{Proceedings of the IEEE conference on computer vision and
  pattern recognition}, pages 770--778, 2016.

\bibitem[Howard et~al.(2017)Howard, Zhu, Chen, Kalenichenko, Wang, Weyand,
  Andreetto, and Adam]{howard2017mobilenets}
Andrew~G Howard, Menglong Zhu, Bo~Chen, Dmitry Kalenichenko, Weijun Wang,
  Tobias Weyand, Marco Andreetto, and Hartwig Adam.
\newblock Mobilenets: Efficient convolutional neural networks for mobile vision
  applications.
\newblock \emph{arXiv preprint arXiv:1704.04861}, 2017.

\bibitem[Hu et~al.(2015)Hu, Yang, and Sim]{hu2015}
Yaohua Hu, Xiaoqi Yang, and Chee-Khian Sim.
\newblock Inexact subgradient methods for quasi-convex optimization problems.
\newblock \emph{European Journal of Operational Research}, 240\penalty0
  (2):\penalty0 315--327, 2015.

\bibitem[Hu et~al.(2016)Hu, Yu, and Li]{hu2016}
Yaohua Hu, Carisa Yu, and Chong Li.
\newblock Stochastic subgradient method for quasi-convex optimization problems.
\newblock \emph{Journal of nonlinear and convex analysis}, 17:\penalty0
  711--724, 01 2016.

\bibitem[Hubara et~al.(2016)Hubara, Courbariaux, Soudry, El-Yaniv, and
  Bengio]{hubara2016binarized}
Itay Hubara, Matthieu Courbariaux, Daniel Soudry, Ran El-Yaniv, and Yoshua
  Bengio.
\newblock Binarized neural networks.
\newblock \emph{Advances in neural information processing systems}, 29, 2016.

\bibitem[Jacob et~al.(2018)Jacob, Kligys, Chen, Zhu, Tang, Howard, Adam, and
  Kalenichenko]{jacob2018quantization}
Benoit Jacob, Skirmantas Kligys, Bo~Chen, Menglong Zhu, Matthew Tang, Andrew
  Howard, Hartwig Adam, and Dmitry Kalenichenko.
\newblock Quantization and training of neural networks for efficient
  integer-arithmetic-only inference.
\newblock In \emph{Proceedings of the IEEE conference on computer vision and
  pattern recognition}, pages 2704--2713, 2018.

\bibitem[Kiwiel and Murty(1996)]{kiwiel1996}
Krzysztof Kiwiel and Katta Murty.
\newblock Convergence of the steepest descent method for minimizing quasiconvex
  functions.
\newblock \emph{Journal of Optimization Theory and Applications}, 89, 04 1996.
\newblock \doi{10.1007/BF02192649}.

\bibitem[Kiwiel(2001)]{kiwiel2001}
Krzysztof~C. Kiwiel.
\newblock Convergence and efficiency of subgradient methods for quasiconvex
  minimization.
\newblock \emph{Mathematical Programming}, 90:\penalty0 1--25, 2001.

\bibitem[Li et~al.(2021)Li, Liu, Yu, Liu, Xu, and Partovi~Nia]{li2021s}
Xinlin Li, Bang Liu, Yaoliang Yu, Wulong Liu, Chunjing Xu, and Vahid
  Partovi~Nia.
\newblock S3: Sign-sparse-shift reparametrization for effective training of
  low-bit shift networks.
\newblock \emph{Advances in Neural Information Processing Systems},
  34:\penalty0 14555--14566, 2021.

\bibitem[Liu et~al.(2018)Liu, Simonyan, and Yang]{liu2018darts}
Hanxiao Liu, Karen Simonyan, and Yiming Yang.
\newblock Darts: Differentiable architecture search.
\newblock \emph{arXiv preprint arXiv:1806.09055}, 2018.

\bibitem[Luo et~al.(2017)Luo, Wu, and Lin]{luo2017thinet}
Jian-Hao Luo, Jianxin Wu, and Weiyao Lin.
\newblock Thinet: A filter level pruning method for deep neural network
  compression.
\newblock In \emph{Proceedings of the IEEE international conference on computer
  vision}, pages 5058--5066, 2017.

\bibitem[Mahajan et~al.(2018)Mahajan, Girshick, Ramanathan, He, Paluri, Li,
  Bharambe, and Van Der~Maaten]{mahajan2018exploring}
Dhruv Mahajan, Ross Girshick, Vignesh Ramanathan, Kaiming He, Manohar Paluri,
  Yixuan Li, Ashwin Bharambe, and Laurens Van Der~Maaten.
\newblock Exploring the limits of weakly supervised pretraining.
\newblock In \emph{Proceedings of the European conference on computer vision
  (ECCV)}, pages 181--196, 2018.

\bibitem[Polyak(1967)]{polyak1967}
Boris Polyak.
\newblock A general method for solving extremum problems.
\newblock \emph{Soviet Mathematics. Doklady}, 8, 01 1967.

\bibitem[Radford et~al.(2019)Radford, Wu, Child, Luan, Amodei, Sutskever,
  et~al.]{radford2019language}
Alec Radford, Jeffrey Wu, Rewon Child, David Luan, Dario Amodei, Ilya
  Sutskever, et~al.
\newblock Language models are unsupervised multitask learners.
\newblock \emph{OpenAI blog}, 1\penalty0 (8):\penalty0 9, 2019.

\bibitem[Ram et~al.(2009)Ram, Nedić, and Veeravalli]{ram2009}
S.~Sundhar Ram, A.~Nedić, and V.~V. Veeravalli.
\newblock Incremental stochastic subgradient algorithms for convex
  optimization.
\newblock \emph{SIAM Journal on Optimization}, 20\penalty0 (2):\penalty0
  691--717, 2009.
\newblock \doi{10.1137/080726380}.
\newblock URL \url{https://doi.org/10.1137/080726380}.

\bibitem[Ramakrishnan et~al.(2020)Ramakrishnan, Sari, and
  Nia]{ramakrishnan2020differentiable}
Ramchalam~Kinattinkara Ramakrishnan, Eyyub Sari, and Vahid~Partovi Nia.
\newblock Differentiable mask for pruning convolutional and recurrent networks.
\newblock In \emph{2020 17th Conference on Computer and Robot Vision (CRV)},
  pages 222--229. IEEE, 2020.

\bibitem[Sanh et~al.(2019)Sanh, Debut, Chaumond, and Wolf]{sanh2019distilbert}
Victor Sanh, Lysandre Debut, Julien Chaumond, and Thomas Wolf.
\newblock Distilbert, a distilled version of bert: smaller, faster, cheaper and
  lighter.
\newblock \emph{arXiv preprint arXiv:1910.01108}, 2019.

\bibitem[Schmidt et~al.(2011)Schmidt, Roux, and Bach]{schmidt2011}
Mark Schmidt, Nicolas Roux, and Francis Bach.
\newblock Convergence rates of inexact proximal-gradient methods for convex
  optimization.
\newblock \emph{Advances in neural information processing systems}, 24, 2011.

\bibitem[Shoeybi et~al.(2019)Shoeybi, Patwary, Puri, LeGresley, Casper, and
  Catanzaro]{shoeybi2019megatron}
Mohammad Shoeybi, Mostofa Patwary, Raul Puri, Patrick LeGresley, Jared Casper,
  and Bryan Catanzaro.
\newblock Megatron-lm: Training multi-billion parameter language models using
  model parallelism.
\newblock \emph{arXiv preprint arXiv:1909.08053}, 2019.

\bibitem[Vaswani et~al.(2017)Vaswani, Shazeer, Parmar, Uszkoreit, Jones, Gomez,
  Kaiser, and Polosukhin]{vaswani2017attention}
Ashish Vaswani, Noam Shazeer, Niki Parmar, Jakob Uszkoreit, Llion Jones,
  Aidan~N Gomez, {\L}ukasz Kaiser, and Illia Polosukhin.
\newblock Attention is all you need.
\newblock \emph{Advances in neural information processing systems}, 30, 2017.

\bibitem[Wu et~al.(2020)Wu, Judd, Zhang, Isaev, and
  Micikevicius]{wu2020integer}
Hao Wu, Patrick Judd, Xiaojie Zhang, Mikhail Isaev, and Paulius Micikevicius.
\newblock Integer quantization for deep learning inference: Principles and
  empirical evaluation.
\newblock \emph{arXiv preprint arXiv:2004.09602}, 2020.

\bibitem[Zhai et~al.(2022)Zhai, Kolesnikov, Houlsby, and
  Beyer]{zhai2022scaling}
Xiaohua Zhai, Alexander Kolesnikov, Neil Houlsby, and Lucas Beyer.
\newblock Scaling vision transformers.
\newblock In \emph{Proceedings of the IEEE/CVF Conference on Computer Vision
  and Pattern Recognition}, pages 12104--12113, 2022.

\bibitem[Zhang et~al.(2022)Zhang, Wilson, and De~Sa]{zhang2022low}
Ruqi Zhang, Andrew~Gordon Wilson, and Christopher De~Sa.
\newblock Low-precision stochastic gradient langevin dynamics.
\newblock In \emph{International Conference on Machine Learning}, pages
  26624--26644. PMLR, 2022.

\bibitem[Zhang et~al.(2020)Zhang, Liu, Zhang, Liu, Huang, Zhou, Guo, Guo, Du,
  Zhi, et~al.]{zhang2020fixed}
Xishan Zhang, Shaoli Liu, Rui Zhang, Chang Liu, Di~Huang, Shiyi Zhou, Jiaming
  Guo, Qi~Guo, Zidong Du, Tian Zhi, et~al.
\newblock Fixed-point back-propagation training.
\newblock In \emph{Proceedings of the IEEE/CVF Conference on Computer Vision
  and Pattern Recognition}, pages 2330--2338, 2020.

\bibitem[Zhao et~al.(2021)Zhao, Huang, Pan, Li, Zhang, Gu, and
  Xu]{zhao2021distribution}
Kang Zhao, Sida Huang, Pan Pan, Yinghan Li, Yingya Zhang, Zhenyu Gu, and
  Yinghui Xu.
\newblock Distribution adaptive int8 quantization for training cnns.
\newblock In \emph{Proceedings of the AAAI Conference on Artificial
  Intelligence}, volume~35, pages 3483--3491, 2021.

\bibitem[Zoph et~al.(2018)Zoph, Vasudevan, Shlens, and Le]{zoph2018learning}
Barret Zoph, Vijay Vasudevan, Jonathon Shlens, and Quoc~V Le.
\newblock Learning transferable architectures for scalable image recognition.
\newblock In \emph{Proceedings of the IEEE conference on computer vision and
  pattern recognition}, pages 8697--8710, 2018.

\end{thebibliography}

\onecolumn
\aistatstitle{Supplementary Materials}
% \aistatsauthor{ Author 1 \And Author 2 \And  Author 3 }

% \aistatsaddress{ Institution 1 \And  Institution 2 \And Institution 3 } 
\section{Proofs}
Let $\x \in \RRn$ and $r \in \RR^+$. Define $\BB(\x,r)=\left\{\x' \in \RRn|\; \|\x'-\w\|\leq r\right\}$. Given that $f$ is quasi-convex and satisfies a Holder condition, and $\X$ is a compact set, it follows that $\Xs\neq \emptyset$ and $f^*>-\infty$. We start with two technical lemmas needed for the proof of Theorem~\ref{theorem1}. 

\begin{lemma} \label{lemma_aux1}
Let $\g \in \RRn$ be a $d$-dimensional vector and $R,C,B,\t \in \RR^+$ be real constants such that $\|\g\|=1$, $R<1$ and $C> B$. Define $F(\x,\r)=  \t\|\g+\r\|^2-2\langle \x,\g+\r\rangle$ where $\r,\w\,\in \RRn$. Consider the following optimization problem,
\begin{align}
    F^* =\max_{\x,\r\in\RRn}\,&F(\x,\r),\;\label{optprob} \\
    \text{subject to:} & \notag\\
       & \langle \g,\x\rangle  \geq B,  \notag  \\
       & \|\r\|^2              \leq R^2, \notag \\
       & \|\x\|^2              \leq C^2.\label{lemma1eqboundW}  
\end{align}

Then 
\[
F^*\leq \max\left\{ \t\left(1+R^2\right)+2R\sqrt{\t^2+C^2-2\t B}-2B,\t\left(R^2-1\right)\right\} .
\]
\end{lemma}
\begin{corollary}
\label{coroptprob}

The upper bound in Lemma~\ref{lemma_aux1} is attained if $d \geq 2$. 
%Moreover if $d \geq 2$, then 
%\[
%F^*=\max\left\{ \t\left(1+R^2\right)+2R\sqrt{\t^2+c^2-2\t B}-2B,\t\left(R^2-1\right)\right\} .
%\]
\end{corollary}

\begin{lemma}{(Lemma 6 of \citet{kiwiel2001})}
\label{lemma_aux2}
Suppose $f(\cdot)$ satisfies an H$\ddot{\text{o}}$lder condition on $\RRn$ with parameter $p$ and constant $L$. 
%fulfills the conditions of Theorem \ref{theorem1}. 
If $\BB(\x^*,M)\subset   \S_{f,f(\x)}, $ then $\langle \x-\x^*,\g/\|\g\|\rangle\geq M, \quad \forall \g \in \bar{\partial}^* f(\x)$.
\end{lemma}

Theorem \ref{theorem1} is proved by contradiction using a standard argument for this type  of results. 
%We show that if the desired result fails to hold, then there is an infinite iterations where the decrease of the  distance from an optimal point is strictly greater than a fixed quantity. 
%
%We remind that the diameter of a compact set is defined as $c=\min\{d\in\RR \,|\,\forall \; \x_1,\x_2 \in \X, \;  \|\x_1-\x_2\|\leq d\}$, thus $\|\x_k-\x^*\|\leq c \; \forall\, k$.
%
For the sake of simplicity, we prove the theorem assuming that $S=0$. The proof for the general case, i.e. $S\neq 0$, is similar upon noticing that $\x_{k+1}=\piX \left[\x_k-\t\left(\frac{\g_k}{\|\g_k\|}+\r_k+\frac{\s_k}{\t}\right)\right]$  and replacing $R$ with $R+\frac{S}{\t}$. Given that $\t$ is assumed to be fixed, this change does not cause any difficulty. 
% since $\t$ is constant in the algorithm.

\subsection{Proof of Theorem \ref{theorem1}}
\begin{proof}

Given that $c$ is the diameter of $\X$, the desired result clearly holds if $\Gamma(c)\geq c$. Thus, assume $\Gamma(c)<c$.
Suppose that $\exists \ \delta > 0, \; \bar{k}>0 \;s.t. \, f(\x_k)>f^*+L(\Gamma(c)+\delta)^p\quad \forall\, k>\bar{k}$ 
and $\x^*\in \X^*$. Given that $f(\cdot)$ satisfies the H$\ddot{\text{o}}$lder condition, we have for all $k>\bar{k}$
\[
f(\x)-f^*\leq L \dist(\x,\X^*)^p \leq L(\Gamma(c)+\delta)^p <  f(\x_k)-f^*, \qquad \forall \, \x \in \BB(\x^*,\Gamma(c)+\delta,) \quad,
\]
which implies that $\BB(\x^*,\Gamma(c)+\delta)\subset   \S_{f,f(\x_k)}, \; \forall\, k>\bar{k}$. Using Lemma~\ref{lemma_aux2}
\begin{equation*}
  \langle \frac{\g_k}{\|\g_k\|},\x_k-\x^*\rangle \geq \Gamma(c) +\delta,\quad\forall\, k>\bar{k},   
\end{equation*}

hence one can use Lemma~\ref{lemma_aux1} with $\;B=\Gamma(c)+\delta, \,C=c, \,\g=\frac{\g_k}{\|\g_k\|}, \,\x= \x^*-\x_k, \,\r=\r_k, \,R=R\,$ and $\t=\t$.  This yields
%   \begin{align*}
%   \langle \x^*-\x,\g_k\rangle+\langle \x^*-\x_,\r_k\rangle+\frac{\t}{2}\|\frac{\g_k}{\|\g_k\|}+\r_k\|^2\leq \max\left\{\frac{\t}{2}\left(1+R^2\right)+R\sqrt{\t^2+c^2-2\t\left[\Gamma(c)+\delta\right]}-\Gamma(c)-\delta,\frac{\t}{2
%   }\left(R^2-1\right)\right\},
% \end{align*}
% that is
   \begin{equation}
   \label{eq:th1updatebound}
   \hspace{-.1 in}\langle \x_k-\x^*,\g_k\rangle+\langle \x_k-\x^*,\r_k\rangle-\frac{\t}{2}\|\frac{\g_k}{\|\g_k\|}+\r_k\|^2\geq \min\left\{\Gamma(c)+\delta-\frac{\t}{2}\left(1+R^2\right)-R\sqrt{\t^2+c^2-2\t\left[\Gamma(c)+\delta\right]},\frac{\t}{2
   }\left(1-R^2\right)\right\}.
\end{equation}
It is shown that for $\delta$ is small enough $$\Gamma(c)+\delta-\frac{\t}{2}\left(1+R^2\right)-R\sqrt{\t^2+c^2-2\t\left[\Gamma(c)+\delta\right]}<\frac{\t}{2
   }\left(1-R^2\right).$$
   Given that $\Gamma(c)$ is the maximum of two terms, we show the inequality holds for both terms, and hence for their maximum. If $\Gamma(c) = \frac{\t}{2}(1+R^2)$, then
\begin{equation}\label{eq:th1diseq1}
    \Gamma(c)+\delta-\frac{\t}{2}\left(1+R^2\right)-R\sqrt{\t^2+c^2-2\t\left[\Gamma(c)+\delta\right]}=\delta-R\sqrt{\t^2+c^2-2\t\left[\Gamma(c)+\delta\right]}\leq0\leq \frac{\t}{2
   }\left(1-R^2\right),
\end{equation}
where the first inequality follows from the fact that we can choose $\delta$ arbitrarily small and second one from $R<1$. 
Otherwise, $\Gamma(c)=\frac{\t(1-R^2)}{2}+Rc$, thus
\begin{alignat}{6}
      \left[\Gamma(c)+\delta-\frac{\t}{2}\left(1+R^2\right)\right]^2
      %&\notag\\
      &=\frac{\t^2}{4}\left(1+R^2\right)^2+\left[\Gamma(c)+\delta\right]^2-\t\left[\Gamma(c)+\delta\right]\left(1+R^2\right)\notag \\
      &=\frac{\t^2}{4}\left(1+R^2\right)^2+\left[\frac{\t\left(1-R^2\right)}{2}+Rc+\delta\right]^2-\t\left[\frac{\t\left(1-R^2\right)}{2}+R\d+\delta\right]\left(1+R^2\right) \notag\\
      &=\frac{\t^2}{4}\left(1+R^2\right)^2+\frac{\t^2\left(1-R^2\right)^2}{4}+R^2c^2+\t Rc\left(1-R^2\right)-\frac{\t^2\left(1-R^4\right)}{2}\notag\\
      &\hspace{.1 in}-\t Rc\left(1+R^2\right)+\delta^2+2\delta Rc+\delta \t \left(1-R^2\right)-\delta \t\left(1+R^2\right)\notag\\
      &= \frac{\t^2}{2}\left(1+R^4\right)+R^2c^2-2\t cR^3-\frac{\t^2\left(1-R^4\right)}{2}+\delta^2+2\delta Rc-2\delta \t R^2\notag\\
      &=R^4\t^2+R^2c^2-2\t cR^3+\delta^2+2\delta Rc-2\delta \t R^2\notag\\
      &=R^2\left(c^2+\t^2R^2-2\t Rc-2\t\delta\right)+\delta^2+2\delta Rc\notag \\
      &=R^2\left[\t^2+c^2-\t^2\left(1-R^2\right)-2\t Rc-2\t\delta\right]+\delta^2+2\delta Rc\notag \\
      &=R^2\left\{\t^2+c^2-2\t\left[\Gamma(c)+\delta\right]\right\}+\delta^2+2\delta Rc.\notag
\end{alignat}

Taking the square root of the first and last term
\begin{align}\label{eq:th1boundfactcomp}
\Gamma(c)+\delta-\frac{\t}{2}\left(1+R^2\right)=\sqrt{R^2\left\{\t^2+c^2-2\t\left[\Gamma(c)+\delta\right]\right\}+\delta^2+2\delta Rc}.
\end{align}
Then, the first term in the $\min$ operator of \REF{eq:th1updatebound} is
\begin{alignat}{6}
\Gamma(c)+\delta-\frac{\t}{2}\left(1+R^2\right)- R\sqrt{\t^2+c^2-2\t\left[\Gamma(c)+\delta\right]}&=\sqrt{R^2\left\{\t^2+c^2-2\t\left[\Gamma(c)+\delta\right]\right\}+\delta^2+2\delta Rc}\\&\hspace{.1 in}- R\sqrt{\t^2+c^2-2\t\left[\Gamma(c)+\delta\right]}\xmapsto{\delta\rightarrow0}0.\notag
\end{alignat}

Hence, if $\delta$ is small enough 
\begin{equation}\label{eq:th1diseq2}
  \Gamma(c)+\delta-\frac{\t}{2}\left(1+R^2\right)- R\sqrt{\t^2+c^2-2\t\left[\Gamma(c)+\delta\right]}\leq \frac{\t}{2}\left(1-R^2\right).  
\end{equation}

If $\delta$ is small enough, Eq.~\REF{eq:th1updatebound}, \REF{eq:th1diseq1} and \REF{eq:th1diseq2} yield
   \begin{equation}
   \label{eq:th1updatebound2}
   \langle \x_k-\x^*,\g_k\rangle+\langle \x_k-\x^*,\r_k\rangle-\frac{\t}{2}\|\frac{\g_k}{\|\g_k\|}+\r_k\|^2\geq \Gamma(c)+\delta-\frac{\t}{2}\left(1+R^2\right)-R\sqrt{\t^2+c^2-2\t\left[\Gamma(c)+\delta\right]}.
\end{equation}

It follows from the argument leading to Eq.~\REF{eq:th1boundfactcomp} that if $\Gamma(c)\geq\frac{\t(1-R^2)}{2}+Rc$,
\begin{equation}\label{eq:th1bounddis}
    \Gamma(c)+\delta-\frac{\t}{2}(1+R^2)\geq\sqrt{R^2\left\{\t^2+c^2-2\t\left[\Gamma(c)+\delta\right]\right\}+\delta^2+2\delta Rc}.
\end{equation}

 Having noted that $\Gamma(c)>\frac{\t}{2} $, we obtain $\t^2+c^2-2\t(\Gamma(c)+\delta)<c^2 $. This yields
\begin{alignat}{6}
R^2\left\{\t^2+c^2-2\t\left[\Gamma(c)+\delta\right]\right\}+\delta^2+2\delta Rc&=\left\{R\sqrt{\t^2+c^2-2\t\left[\Gamma(c)+\delta\right]}+\delta\right\}^2\notag\\ &\hspace{.1 in}-2\delta R\sqrt{\t^2+c^2-2\t\left[\Gamma(c)+\delta\right]}+2R\delta c\label{eq:th1bounddis2}\\ &\geq \left\{R\sqrt{\t^2+c^2-2\t\left[\Gamma(c)+\delta\right]}+\delta\right\}^2\notag. 
\end{alignat}

It follows from Eq.~\REF{eq:th1updatebound2}, \REF{eq:th1bounddis} and \REF{eq:th1bounddis2} that

\begin{equation*}
 \langle \x_k-\x^*,\g_k\rangle+\langle \x_k-\x^*,\r_k\rangle-\frac{\t}{2}\|\frac{\g_k}{\|\g_k\|}+\r_k\|^2\geq \delta,
\end{equation*}

leading to \begin{equation*}
\|\x_{k+1}-\x^*\|\leq
\|\x_k-\x^*\|^2-2\t\left({\langle \x_k-\x^*,\g_k\rangle+\langle \x_k-\x^*,\r_k\rangle-\frac{\t}{2}\|\frac{\g_k}{\|\g_k\|}+\r_k\|^2}\right)\leq
\|\x_k-\x^*\|^2-2\t \delta,
\end{equation*}
where we used the definition of $\x_{k+1}$ and the property of the projection map $\piX(\cdot)$ in the first inequality.

Thus
\begin{equation}\label{eq:th1final}
    \|\x_{k+1}-\x^*\|^2\leq\|\x_k-\x^*\|-2\t \delta\leq \ldots\leq \|\x_0-\x^*\|-2\t(k-\bar{k}+1)\delta,\quad \forall\, k>\bar{k}.
\end{equation}
 
 The desired result then follows upon noticing that the upper bound tends to $-\infty$ as $k$ tends to $+\infty$ which leads to a contradiction.

\end{proof}

\subsubsection{Intuition behind Lemma \ref{lemma_aux1}}

By scrutinizing the proof of our Theorem~\ref{theorem1} and Theorem 3.1 of \citep{hu2015}, we notice that the key to the proof is
% \[
% \langle \x_k-\x^*,\frac{\g_k}{\|\g_k\|}\rangle>B+\delta>0
% \]with $B>0$. This yields to
\begin{align*}
    \|\x_{k+1}-\x^*\|^2&\leq\|\x_k-t\hat{\g}_k-\x^*\|^2=\|\x_k-\x^*\|^2-2\t\langle \x_k-\x^*,\frac{\g_k}{\|\g_k\|}\rangle-2\t\langle \x_k-\x^*,\r_k\rangle+\t^2\|\frac{\g_k}{\|\g_k\|}+\r_k\|^2\\
&=\|\x_k-\x^*\|^2-2\t\left({B+\langle \x_k-\x^*,\r_k\rangle-\frac{\t}{2}\|\frac{\g_k}{\|\g_k\|}+\r_k\|^2}\right),
\end{align*}
where $B$ should be such that $\exists \delta'>0\;|\;\left({B+\langle \x_k-\x^*,\r_k\rangle-\frac{\t}{2}\|\frac{\g_k}{\|\g_k\|}+\r_k\|^2}\right)>\delta'$. The final bound will significantly depend on $B$. More precisely, greater $B$ values lead to greater, hence worse, bounds.

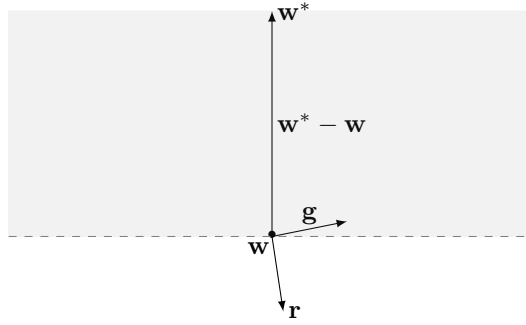
\begin{figure}[!ht]
    \centering
  \begin{tikzpicture}[scale=1,inner sep=0pt, outer sep=2pt,>=latex]
\draw[->](0,0)--(1,0.2)node[midway, above]{$\g$}; 
\draw[->](0,0)--(0,3)node[midway,right]{$\x^*-\x$};
\draw[->](0,0)--(0.15,-1)node[right]{$\r$};
\draw(0,0)node{•}node[below,xshift=-5]{$\x$};
\draw[dashed, color=gray](-3.5,0)--(3.5,0);
\draw(0,3)node[right]{$\x^*$};
\draw[draw=gray,fill=gray, opacity=0.1] (-3.5,0) rectangle (3.5,3);
\end{tikzpicture}
    \caption{Given $\x$ and $\x^*$, $\g$ needs to have positive scalar product with $\x^*-\x$ (gray zone). So it s not possible for $\r$ to be opposite to $\x^*-\x$ and colinear with $\g$ at the same time.}
    \label{fig:tkz1}
\end{figure}

\begin{figure}[!ht]
    \centering
  \begin{tikzpicture}[scale=1,inner sep=0pt, outer sep=2pt,>=latex]
\draw[->](0,0)--(1,0.2)node[midway, above]{$\t\g$}; 
\draw[->, color=gray](1,0.2)--(0,3)node[right, midway]{$\x^*-(\x+\t\g)$}; 
\draw[->, color=red, dashed](1,0.2)--(1.5,-1.2)node[right, midway, color=red]{$\r$};
\draw[->](0,0)--(0,3)node[midway,left]{$\x^*-\x$};
\draw[->,color=red](0,0)--(0.5,-1.4)node[right]{$\r$};
\draw(0,0)node{•}node[below,xshift=-5]{$\x$};
\draw[dashed, color=gray](-3.5,0)--(3.5,0);
\draw(0,3)node[right]{$\x^*$};
\draw[draw=gray,fill=gray, opacity=0.1] (-3.5,0) rectangle (3.5,3);
\end{tikzpicture}
    \caption{Given $\x$, $\x^*$ and $\g$ the worst case seems to be $-\r=\x^*-(\x+\t\g)$, scaled to have norm $R$.}
    \label{fig:tkz2}
\end{figure}
In \citet{hu2015} the authors use

\begin{equation}
\label{hueq1}
    |\x_k-\x^*\|^2-2\t\left(B+\langle \x_k-\x^*,\r_k\rangle-\frac{\t}{2}\|\frac{\g_k}{\|\g_k\|}+\r_k\|^2\right)\leq \|\x_k-\x^*\|^2-2\t\left(B-Rc-\frac{\t}{2}(1+R)^2\right),
\end{equation}
\[
\
\]
and then choose $B=Rc+\frac{\t}{2}(1+R)^2+\delta$. To obtain Eq.~\REF{hueq1}, \citet{hu2015} uses
\begin{align}
    \langle \x_k-\x^*,\r_k\rangle\geq -\|\x_k-\x^*\|\|\r_k\|,\label{eq:inshhu1}\\
    \|\frac{\g_k}{\|\g_k\|}+\r_k\|\leq \|\frac{\g_k}{\|\g_k\|}\|+\|\r_k\|.\label{eq:inshhu2}
\end{align}

The equality holds in \REF{eq:inshhu1} and \REF{eq:inshhu2} if $\r_k = -\alpha (\x_k-\x^*)$ and $\r_k = \beta \g_k$ with $\alpha,\;\beta>0$, respectively. This then implies that $\langle \g_k, \x_k-\x^*\rangle = \langle \r_k/\beta, -\r_k/  \alpha\rangle=-\frac{1}{\alpha\beta}\|\r_k\|^2<0$, which contradicts the definition of quasi sub-differential for $\g_k\in \bar\partial^*f(\x_k)$.
This means that equality cannot hold in Eq.~\REF{eq:inshhu1} and \REF{eq:inshhu2} simultaneously.  That is why, we used Lemma~\ref{lemma_aux1} instead of Eq.~\REF{hueq1} to prove Theorem~\ref{theorem1}. Figure~\ref{fig:tkz1} provides the insight for our argument.

Figure~\ref{fig:tkz2} shows that the worst case scenario happens if $\r_k$ is collinear with  $\x_k+\t\frac{\g_k}{\|\g_k\|}-\x^*$. The proof is given in Section~\ref{lemmasproof}.

\subsubsection{Proof of Lemma~\ref{lemma_aux1}}
% The scheme of the proof is not to find the optimal solutions of Problem~\ref{optprob}.
Loosely speaking, the idea is to use KKT's conditions and to prove that the optimal solutions are on the boundary of the feasible set. 

\begin{proof}
\label{lemmasproof}
The gradient of the objective function is 
\[
\nabla F(\cdot)= (-2(\g+\r),2\t(\g+\r)-2\x).
\]
The objective function is convex so the optimum is reached in at least a point on the boundary of the feasible region. 
We recall that the weak Slater's conditions for Problem~\ref{optprob} are satisfied if $\exists \, \r,\x \in \RRn$ such that
\begin{align*}
\langle \g,\x\rangle\geq B,\\
\|\r\|^2<R^2,\\
\|\x\|^2<C^2.
\end{align*}
These conditions are fulfilled by $\r=0$ and $\w=B\g$, hence we can use the KKT's equations to obtain necessary conditions on the optimal solutions of Problem~\REF{optprob}. From the KKT's condition on the gradient component with respect to $\r$
\begin{alignat}{6}
    \frac{\partial F}{\partial \r} = \lambda \frac{\partial(\|\r\|^2-R^2) }{\partial \r}  \notag &\Leftrightarrow
    2\t(\g+\r)-2\x = \lambda 2\r\notag 
\end{alignat}
with $\lambda \in \RR^+$.

If  $\t=\lambda$, then $\x=\t\g$, and thus the objective function is simplified to $ \t+\t\|\r\|^2-2\t$. In this case, any $\r$ such that $\|\r\|=R$ achieves the optimal value, being
$\t(R^2-1)$. 
Otherwise,
\begin{equation}
     \r = \frac{\x-\t \g}{\t-\lambda}. \label{eq:lemma1eq1}
\end{equation}
The objective function can be written as
\[
F(\r,\w)=\t+2\langle \t\g-\x,\r\rangle+\t\|\r\|^2-2\langle \x,\g\rangle.
\]

Let $(\r^*,\x^*)$  be an optimal solution of Problem~\REF{optprob}. We prove the following facts by contradiction:

\begin{itemize}
\item[a)]$\langle \r^*,\t\g-\x^*\rangle\geq0.$
    
    Assume $\langle \r^*,\t\g-\x^*\rangle<0$. Consider $\tilde \r =-\r^*$, the solution $(\tilde \r,\x^*)$ is feasible since $\|\tilde \r\|=\|\r^*\|\leq R$. Moreover, $\langle \t\g-\x^*,\tilde \r\rangle>0>\langle \t\g-\x^*,\r^*\rangle $, this yields $$F(\r^*,\x^*)-F(\tilde\r,\x^*)=\langle \t\g-\x^*,\r^*\rangle-\langle \t\g-\x^*,\tilde \r\rangle<0.$$ The last inequality violates the optimality of $\r^*$, therefore the claim is proved.

    \item[b)] $\|\r^*\|=R.$
  
   Assume $\|\r^*\|<R$. Consider $\tilde \r= R\frac{\r^*}{\|\r^*\|}$, using point (a) we have
    $$\langle \t\g-\x^*,\tilde \r\rangle-\langle \t\g-\x^*,r^*\rangle=\left(\frac{R}{\|\r^*\|}-1\right)\langle \t\g-\x^*,\r^*\rangle\geq0.$$ 
   Moreover, $R=\|\tilde \r\|>\|\r^*\|$, hence $$F(\r^*,\x^*)-F(\tilde\r,\x^*)=2\left[\left(\frac{R}{\|\r^*\|}-1\right)\langle \t\g-\x^*,\r^*\rangle\right]+\t(\|\r^*\|-R)<0.$$ 
    The last inequality violates the optimality of $\r^*$ and hence the result follows.
    
    \item[c)]$\langle \x^*,\g\rangle=B.$
    
    Assume $\langle \x^*,\g\rangle>B$. Then $\tilde \x=\x^*-(\langle \x^*,\g\rangle-B)\g$, $(\r^*,\tilde\x)$ is a feasible point since \begin{alignat*}{6}
    \|\tilde\x\|^2=\|\x^*\|^2+(\langle \x^*,\g\rangle-B)^2-2\left[\langle \x^*,\g\rangle-B\right]\langle \x^*,\g\rangle<\|\x^*\|^2
    \end{alignat*} and $\langle \tilde\x,\g\rangle=B$.
    \begin{alignat*}{6}
         F(\x^*,\r^*)-F(\tilde \x,\r^*)=&2\langle \tilde \x-\x^*,\g+\r^*\rangle=-2\langle \left[\langle \x^*,\g\rangle-B\right]\g,\g+\r^*\rangle\\=&-2(\langle \x^*,\g\rangle-B)-2\langle [\langle \x^*,\g\rangle-B]\g,\r^*\rangle\\\leq& -2[\langle \x^*,\g\rangle-B](1-\|\g\|\|\r^*\|)<0.
    \end{alignat*}
   The last inequality violates the optimality of $\x^*$, this completes the proof of this part.
      \end{itemize}
    
    It follows from Eq.~\REF{eq:lemma1eq1}, (a) and (b) that \begin{equation}\label{rstar}
        \r^*=\alpha (\t\g-\x^*)\; \text{ with }\; \alpha=\frac{R}{\|\t\g-\x^*\|}.
    \end{equation}
The desired result follows from \REF{rstar}, \REF{lemma1eqboundW} ,  \begin{align*}
        F(\x^*,\r^*)&= \t(1+R^2)+2R\sqrt{\t^2+\|\w\|^2-2\t B}-2B\\
        &\leq\t(1+R^2)+2R\sqrt{\t^2+c^2-2\t B}-2B.
    \end{align*}
    \end{proof}
    We now present the proof of Corollary~\ref{coroptprob}
    \begin{proof}
    
     Assume $d\geq 2$ and $\|\x^*\|<C$. Consider $\y$ s.t. $\langle \g,\y\rangle=0$ and $\langle \x^*,\y\rangle\geq0$. Since $d\geq2$ we can assume, without loss of generality, that $\|\y\|=1$ . Furthermore, define $\tilde \x= \x^*+a \y$ with $a\in\RR$, then $\exists\;a>0\;s.t.\;\|\tilde \x\| = C$. The point $(\r^*,\tilde \x)$ is a feasible point since $\langle\tilde\x,\g \rangle=\langle \x^*,\g \rangle= B$.

    We consider two cases:
    
    - \underline{$\g$ is not collinear with $\x^*$}.
    
    In this case $\exists\, \y$ such that $\langle \x^*,\y\rangle>0$. Eq.~\REF{rstar} yields
    \begin{align*}
        F(\x^*,\r^*)-F(\tilde \x, \r^*)&=2\langle \tilde \x-\x^*,\g+\r^*\rangle\\&=2\langle a \y,\g+\r^*\rangle\\&=2\langle a \y,\r^*\rangle\\&=2\langle a \y,\alpha [\t\g-\x^*]\rangle\\&=-2a\alpha\langle \y,\x^*\rangle\\&<0.
    \end{align*}
     The last equation violates the optimality of $\x^*$, thus
    \begin{equation}
    \label{lemma1optW}
     \|\w^*\|=C.
 \end{equation} 
    
    - \underline{$\g$ is collinear with $\x^*$}. 
    
    In this case (c) implies $\x^*=B\g$ . Consider $\tilde \r=\beta [(\t-B)\g-a\y]$, then
    \begin{alignat*}{6}
        F(\x^*,\r^*)-F(\tilde \x,\tilde \r)=&2\left[\langle \t\g-\x^*,\r^* \rangle-\langle \t \g-\tilde\x,\tilde\r \rangle-\langle\g,\x^*-\tilde\x \rangle\right]\\
       =&2\left\{\langle(\t-B)\g,\sgn(\t-B)R\g \rangle-\langle(\t-B)\g-a\y,\beta \left[(\t-B)\g-a\y \right] \rangle\right\}\\
        =&2\left\{\langle\sgn(\t-B)(\t-B)\g,R\g \rangle-\beta\langle(\t-B)\g-a\y, (\t-B)\g-a\y  \rangle\right\}\\
        =&2\left\{|\t-B|R-\beta\left[(\t-B)^2+a^2\right]\right\}\\
        =&2\left\{|\t-B|R-R\sqrt{(\t-B)^2+a^2}\right\}\\
        =&R\sqrt{(\t-B)^2+a^2}\left(\frac{|\t-B|}{\sqrt{(\t-B)^2+a^2}}-1\right)\\
        <&0,
    \end{alignat*}
  where the second equality follows from $\x^*=B\g$, Eq.~\REF{rstar} and $\langle \g, \y\rangle=0$, and the fourth and fifth equalities are implied by $\langle \g, \y\rangle=0$.
  The last equation violates the optimality of $(\x^*,\r^*)$, thus \REF{lemma1optW} must be fulfilled.
  The desired result then follows,
    \begin{equation*}
    F(\x^*,\r^*)= \t(1+R^2)+2R\sqrt{\t^2+c^2-2\t B}-2B.
    \end{equation*}
\end{proof}

\subsection{Corollary 4.1.1}
\begin{proof}

Minimizing the bound in Theorem~\ref{theorem1} is equivalent to minimizing
$$\max\left\{\frac{\t}{2}\left[1+\left(R+\frac{S}{\t}\right)^2\right], \frac{\t}{2}\left[1-\left(R+\frac{S}{\t}\right)^2\right]+c \left[ R +\frac{S}{\t}\right]\right\}.$$
Define the function $G(\t)$ as
\begin{align*}
        G(\t):=&\max\left\{\frac{\t}{2}(1+R^2+\frac{S^2}{\t^2}+\frac{2SR}{\t}),\frac{\t(1-R^2-\frac{S^2}{\t^2}-\frac{2RS}{\t})}{2}+Rc+\frac{Sc}{\t}\right\}\\=&
        \max\left\{\frac{\t}{2}(1+R^2)+\frac{S^2}{2\t}+SR,\frac{\t}{2}(1-R^2)-\frac{S^2}{2\t}-RS+Rc+\frac{Sc}{\t}\right\}.
\end{align*}
Note that
\begin{align*}
    G(\t)=\begin{cases}
    \frac{\t}{2}(1+R^2)+\frac{S^2}{2\t}+SR & \text{if }t\geq \frac{c-S}{R}=\t_3,\\
    \frac{\t}{2}(1-R^2)-\frac{S^2}{2\t}-RS+Rc+\frac{Sc}{\t} &\text{otherwise}.
    \end{cases}
\end{align*}
The minimum of the first part is achieved at $\t_1=\frac{S}{\sqrt{1+R^2}}$
while the minimum of the second part is achieved at $\t_2=\sqrt{\frac{S(c-2S)}{1-R^2}}$.
Since $G(\t)$ is the maximum of two convex functions, it attains its global minimum in the set $\left\{\t_1.\t_2,\t_3\right\}$.

\end{proof}

\subsubsection{Proof of Theorem \ref{theorem2}}

The theorem follows immediately from Eq.~\REF{eq:th1final} if $\x_{k+1}=\piX(\x_k-\t\hat{\g}_k+\s_k).$ However, the key to prove Theorem~\ref{theorem2} is $\|\x_{k+1}-\x^*\|\leq\|\x_k-\x^*\|, \; \forall \,0<k<\tilde k$ where $\tilde k=\inf\left\{k|\, f(\x_k)>f^*+L\left[\Gamma(C_0)\right]^p\right\}$. It therefore suffices to establish this last inequality to complete the proof of Theorem~\ref{theorem2}.
\begin{proof}
Fix $K>0$. Suppose $\delta>0$ and $ f(\x_k)>f^*+L(\Gamma(C_0)+\delta)^p, \quad \forall\; k\leq K.$
It is shown by induction that \[\|\x_k-\x^*\|\leq C_0-2\t\delta k, \quad \forall \, k\leq K.\] The case $k=0$ is trivial.

 Suppose $\|\x_j-\x^*\|\leq C_0-2\t\delta j\leq C_0, \; \forall j\leq k$, then the argument of the proof Theorem~\ref{theorem1} can be repeated with $c=C_0$. This is possible because the projection operator in the definition of $\x_{k+1}$ of Theorem~\ref{theorem1} was needed only to bound the norm of $\|\x_k-\x^*\|$ when Lemma~\ref{lemma_aux1} is used. Using a similar argument leading to Eq.~\REF{eq:th1final} and the induction hypothesis, we have
%  from the definition of the algorithm we get

% \[
% \|\x_{k+1}-\x^*\|^2=\|\x_k-\t\hat{\g}_k+s_k-\x^*\|^2=\|\x_k-\x^*\|^2-2\t\langle \x_k-\x^*,\frac{\g_k}{\|\g_k\|}\rangle+2\t\|\x_k-\x^*\|\|\r_k+\frac{s_k}{\t}\|+\t^2\|\frac{\g_k}{\|\g_k\|}+\r_k+\frac{s_k}{\t}\|^2\leq
% \]

% \[
% \|\x_k-\x^*\|^2-2\t(\langle \x_k-\x^*,\frac{\g_k}{\|\g_k\|}\rangle-(R+\frac{S}{\t})\|\x_k-\x^*\|-\frac{\t}{2}(1+R+\frac{S}{\t})^2)
% \]
% Since $\|\x_k-\x^*\|\leq C_0-2\t k\delta<C_0$, we can use \REF{th8eq1} as in the proof of Theorem \ref{theorem1} (with $C=C_0$) and  we obtain
\begin{equation*}
  \|\x_{k+1}-\x^*\|^2\leq\|\x_k-\x^*\|-2\t\delta\leq C_0-2\t\delta(k+1),
\end{equation*}
If $\delta\geq\frac{C_0}{2\t K}$,
the above inequality leads to 
a contradiction since $f(\x_K)>f^*$. Thus $\exists\,k\leq K\;s.t.\;f(\x_k)\leq f^*+L(\Gamma(C_0)+\delta)^p $. This completes the proof.
\end{proof}

\subsubsection{Proof of Corollary 4.3.1}
\begin{proof}
Define $G(\t)=\t(\V+1)+\frac{\W}{\t}$. The function
$G(\cdot)$ is convex and continuosly differentiable in $\RR^+\setminus \{0\}$. Hence the minimum of $G(\cdot)$ is achieved at the roots of its derivative. Obviously only positive roots are of interest.
The derivative of $G(\cdot)$ is
\[
\frac{\partial G(\cdot)}{\partial \t}=\V+1-\frac{\W}{\t^2},
\]
with positive root $\t^*=\sqrt{\frac{\W}{\V+1}}$.

\end{proof}

\subsection{Proof of Theorem \ref{theorem3}}
\begin{proof}
Let $\F_n=\sigma\{\x_0,\x_1,\dots,\x_n\}$ be the $\sigma$-algebra generated  by $\{\x_0,\x_1,\dots,\x_n\}$. Suppose $\x_n \notin \X^*$, it follows from the definition of $\x_{n+1}$
\begin{alignat*}{6}
%\begin{split}
  \|\x_{n+1}-\x^*\|^2  &\leq
  \|\x_n-\t\hat{\g}(\x_n)-\t \r_n-\s_n-\x^*\|^2 \\
  &\leq\|\x_n-\x^*\|^2-2\t\langle \hat{\g}(\x_n),\x_n-\x^*\rangle-2\t\langle \x_n-\x^*,\r_n\rangle \\ &\hspace{.1 in}-2\langle \x_n-\x^*,\s_n\rangle+\t^2\|\hat{\g}(\x_n)+\r_n+\frac{\s_n}{\t}\|^2.
%\end{split}
\end{alignat*}
Taking conditional expectation given $\F_n$
\begin{align}
%\begin{split}
\EE\left\{\|\x_{n+1}-\x^*\|^2|\F_n\right\} %&\leq \notag \\
&\leq   
 \|\x_n-\x^*\|^2 
- 
 2\t\EE\left\{\langle \hat{\g}(\x_n),\x_n-\x^*\rangle|\F_n\right\} 
- 
 2\t\EE\left\{\langle \x_n-\x^*,\r_n\rangle|\F_n\right\}  \notag
 \\
& \hspace{.1 in} - 
2\EE\left\{\langle \x_n-\x^*,\s_n\rangle|\F_n\right\} 
+
\t^2\EE\left\{\|\hat{\g}(\x_n)+\r_n+\frac{\s_n}{\t}\|^2|\F_n\right\} & \notag
\\
&\leq 
 \|\x_n-\x^*\|^2 - 2\t\left(\frac{f(\x)-f^*}{L} \right)^{\frac{1}{p}} - 2\t\langle \x_n-\x^*,\EE\left\{\r_n\right\}\rangle & \notag
 \\
& \hspace{.1 in} -
 2\langle \x_n-x^*,\EE\left\{\s_n\right\}\rangle+\t^2\left(1+\EE\left\{\|\r_n\|^2\right\}+2\EE\left\{\langle\hat{\g}(\x_n),\r_n\rangle|\F_n\right\}\right) &
 \notag 
 \\
& \hspace{.1 in} +
 \EE\left\{\|\s_n\|^2\right\}+2\t\EE\left\{\langle\hat{\g}(\x_n),\s_n\rangle|\F_n\right\}+2\t\EE\left\{\langle \s_n,\r_n\rangle\right\} \notag 
 \\
& \leq 
%\end{split}
\label{th3eq3}
     \|\x_n-\x^*\|^2-2\t\left(\frac{f(\x)-f^*}{L} \right)^{\frac{1}{p}}+\t^2\left(1+\V\right)+\W,
\end{align}
where Lemma~\ref{lemma2} is used to derive the second inequality. The third inequality follows from the ortogonality of $\g(\x_n),\, \r_n,\, \s_n$ and the fact that $\EE\left\{\r_n\right\}=\EE\left\{\s_n\right\}=0$.
Consider now $\delta>0$ and 
\[
\X_{\delta}=\X\cap \S_{f,a},
\]
with $a=f^*+L\left[\frac{\t}{2}\left(1+\V\right)+\frac{\W}{2\t}+\delta\right]^p$.  Since $f$ is continuos, $\exists \y \in \RRn |\;\y_{\delta}\in \X$ and $f(\y_{\delta})=f^*+L(\delta)^p$.
Define the process
\[
\tilde\x_{k+1}=
\begin{cases}
\piX\left[\t\x_k-\t\hat{\g}\left(\tilde \x_k\right)\right],  &\text{if  } \tilde \x_k\notin \X_{\delta},\\
\y_{\delta} &\text{otherwise.}
\end{cases}
\]
We show that $\tilde \x_k \notin \X_{\delta}, \, \forall\, k>K,\,K\in \NN$ leads to a contradiction. Without loss of generality we consider $K=0$.
Assume $\tilde \x_k \notin \X_{\delta}$ for any $k$ and let $\hat{\F}_k=\{\tilde \x_0,\tilde \x_1,\dots,\tilde \x_k\}$. Since $f(\tilde \x_k)\geq f^*+L\left[\frac{\t}{2}\left(1+\V\right)+\frac{\W}{2\t}+\delta\right]^p$, using Eq.~\REF{th3eq3} it follows that $\forall\, \x^*\in \X^*$,
\begin{align*}
%\begin{split}
\EE\left\{\|\tilde \x_{k+1}-\x^*\|^2|\,\hat{\F}_k\right\} %&\leq\\
&\leq  
\|\tilde \x_k - \x^*\|^2 
-2\t \left[\left(\frac{f(\tilde \x_k)-f^*}{L} \right)^{\frac{1}{p}}-\frac{t}{2}(1+\V)-\frac{\W}{2\t} \right] \\ %\leq \\ 
 &\leq  
  \|\tilde \x_k-\x^*\|^2-2\t\delta .
%\end{split}
\end{align*}
 Theorem~\ref{lemma3} then implies $\sum_{k=0}^{+\infty}2\t\delta<+\infty$. This is a contradiction and hence the proof is complete. 
\end{proof}

\begin{lemma}{(Lemma 2.4 of \citet{hu2016}).}\label{lemma2}
Let be $\x \notin \X^*$ and $\hat{\g}(\x)$ a unit noisy quasi sub-gradient of $f$ at $\x$. Then, $\forall \, \x^*\in \X^*$ it holds that, given any $\F=\sigma\{\x_1,...,\x_k\}$ where $\x_i $ are random variables $\forall i=1,..,k$, then
\begin{equation*}
 \EE\left\{\langle \hat{g}(\x),\x^*-\x\rangle|\F\right\}\geq \left (\frac{f(\x)-f^*}{L} \right)^{\frac{1}{p}}.
\end{equation*}
\end{lemma}

\begin{lemma}{(Lemma 2.5 of \citet{hu2016}).}\label{lemma3}
Let $\{Y_k\}$, $\{Z_k\}$ and $\{W_k\}$ be three sequences of nonnegative random, and let $\{\F_k\}$ be a filtration. Suppose that the following conditions are satisfied for each $k$:
\begin{itemize}
\item[a)] $Y_k$, $Z_k$ and $W_k$ are functions of the random variables in $\F_k$;
\item[b)] $\EE\left\{Y_{k+1}|\F_k\right\}\leq Y_k-Z_k+W_k$;
\item[c)] $\sum_{k=0}^{\infty}W_k<+\infty$.
\end{itemize}
Then $\sum_{k=0}^{\infty}Z_k<+\infty$, and the sequence $\{Y_k\}$ converges to a nonnegative random variable $Y$, almost surely.
\end{lemma}

% \begin{lemma}\label{lemma4}
% Let $\x$ and $\x'$ be defined as in Definition~\ref{definition1}

% If $\,\mathcal{P} \left \{\S_{f,f(\x)} \cap \mathcal{A}_{\x} \ne \emptyset \right \} = 0, $
% where $\mathcal{A}_{\x} = \{ \x' : \langle  \g(\x), \x' - \x \rangle >  0 \}, $ then 
% \begin{equation*}
%      \EE\left\{\langle  g(\x), \x'-\x\rangle | \F \right\}\leq 0 \quad \forall \, \x'\in \S_{f,f(\x)}
% \end{equation*}
% \end{lemma}
\section{Details of the experiments setting}
We used the Normalized Gradient Descent (NGD) algorithm to perform the experiments with the deterministic function $f(\x)=\|\x\|^{0.2}$ presented in Section~\ref{sec:experiments}. The maximum number of epochs is set to 1500. Different values for $C_0$, $B_r$, $B_s$ and learning rates were used.
The errors, which  are manually added, have uniform distribution in each coordinate.  Thus, the variances required in Theorem~\ref{theorem3} are $\sigma^2_r= \frac{B_r^2}{3}$ and $\sigma^2_s= \frac{B_s^2}{3}$.
 For the computation of the bound in Theorem~\ref{theorem1}, $c=C_0$ was used.

\subsection{Testing the optimal learning rate}
We performed experiments with fixed values for $B_s$, $B_r$, but different choices of $\t$ to acquire the optimal choice $\t^*$ given by Corollary~\ref{corollary2} . The experiment is repeated 10 times, for each tested value of $\t$,. Finally, the maximum loss function value observed across all the experiments with the same $\t$ is plotted at each epoch. 

\begin{figure}[!ht]\centering      \includegraphics[width=0.465\linewidth]{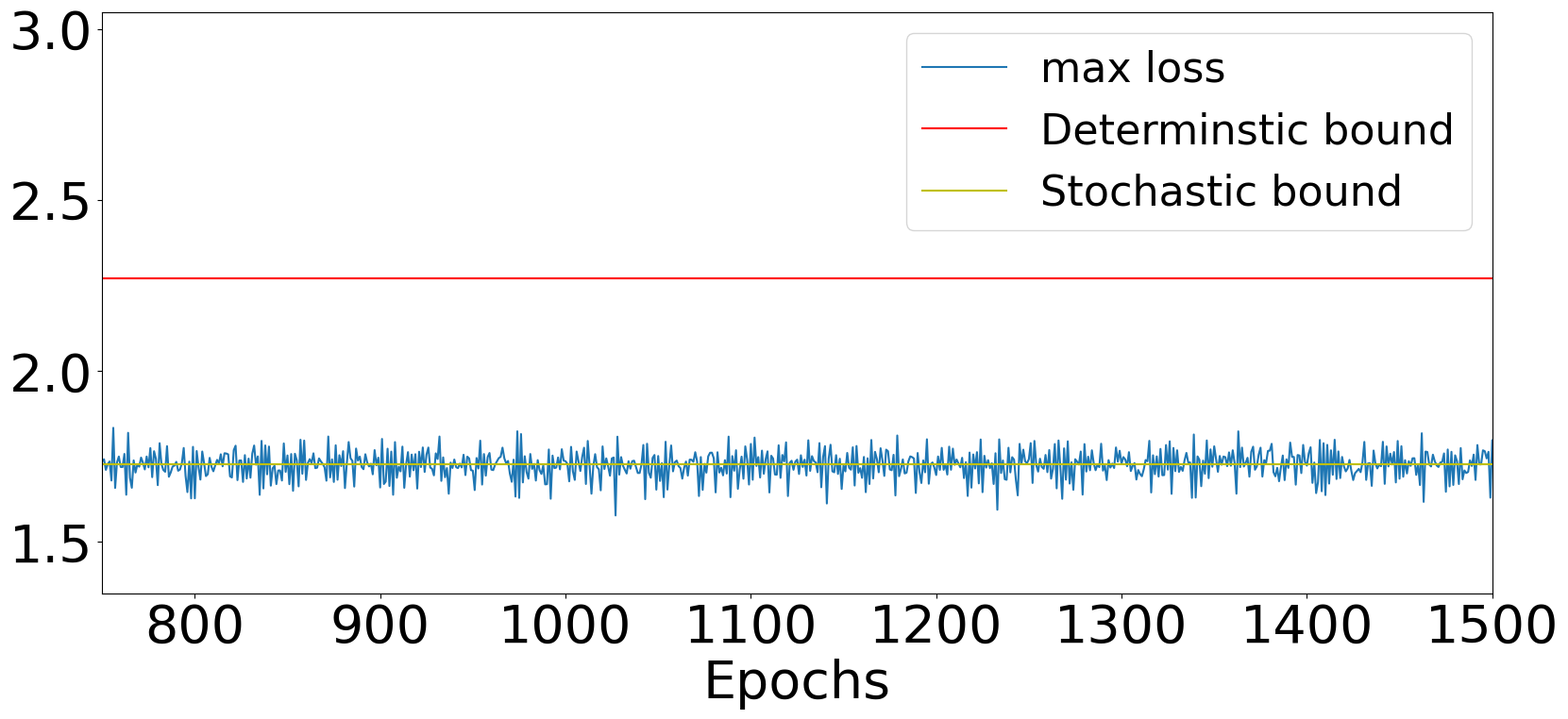}\hspace{.2 in}
      \includegraphics[width=0.465\linewidth]{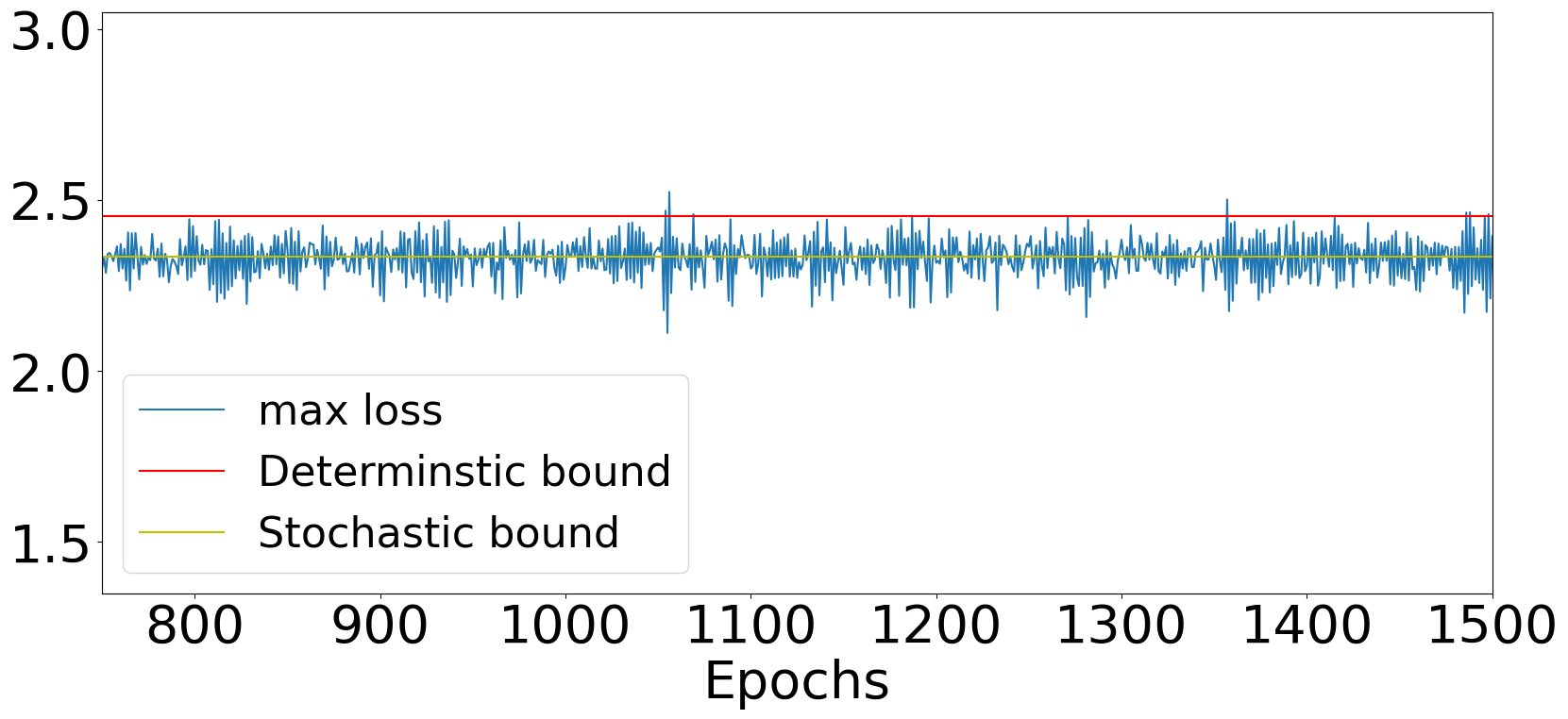}
    	\caption{Results with $\t=0.1$ (left panel) and $\t=0.5$ (right panel)}
    	\label{fig:avg1}
    \end{figure}

\begin{figure}[!ht]
\centering        \includegraphics[width=0.465\linewidth]{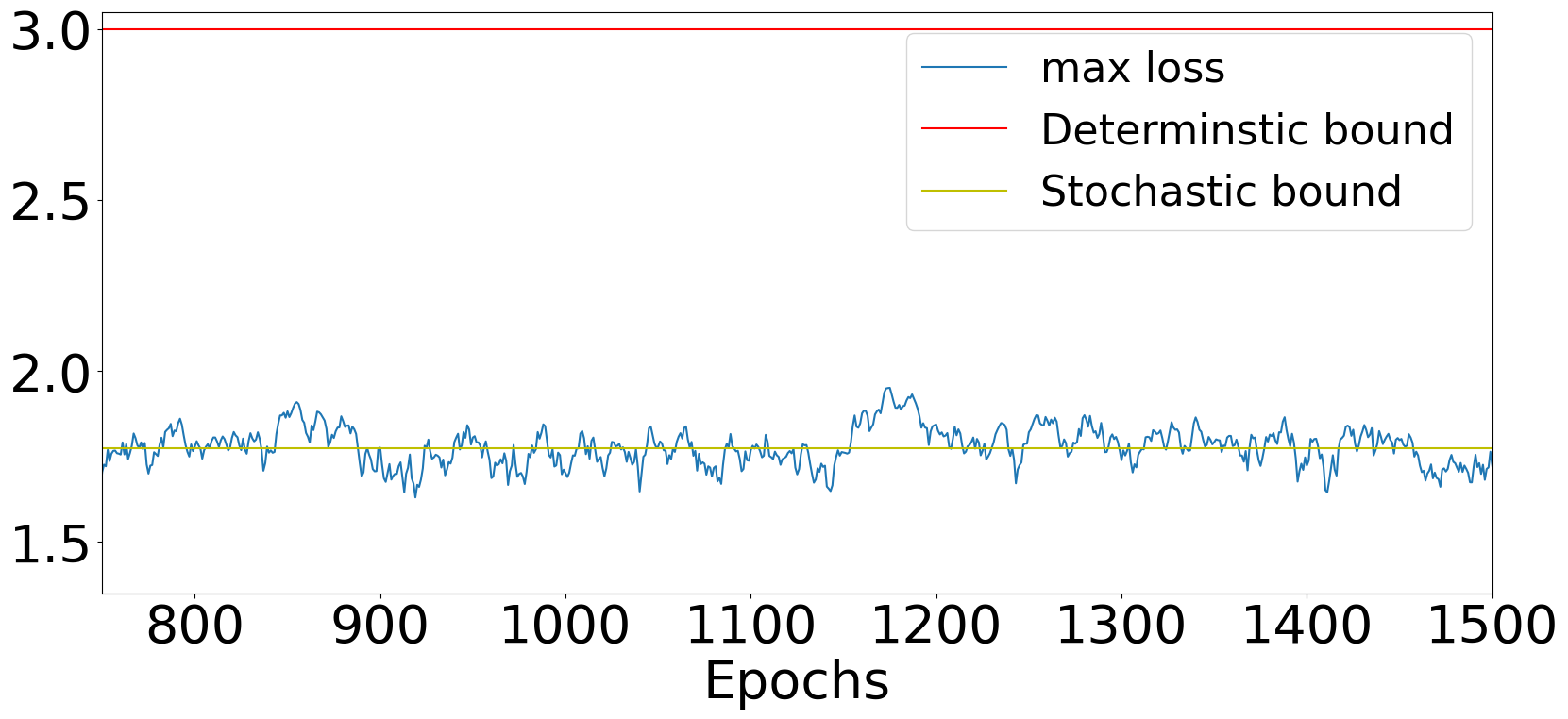}\hspace{.2 in}\includegraphics[width=0.465\linewidth]{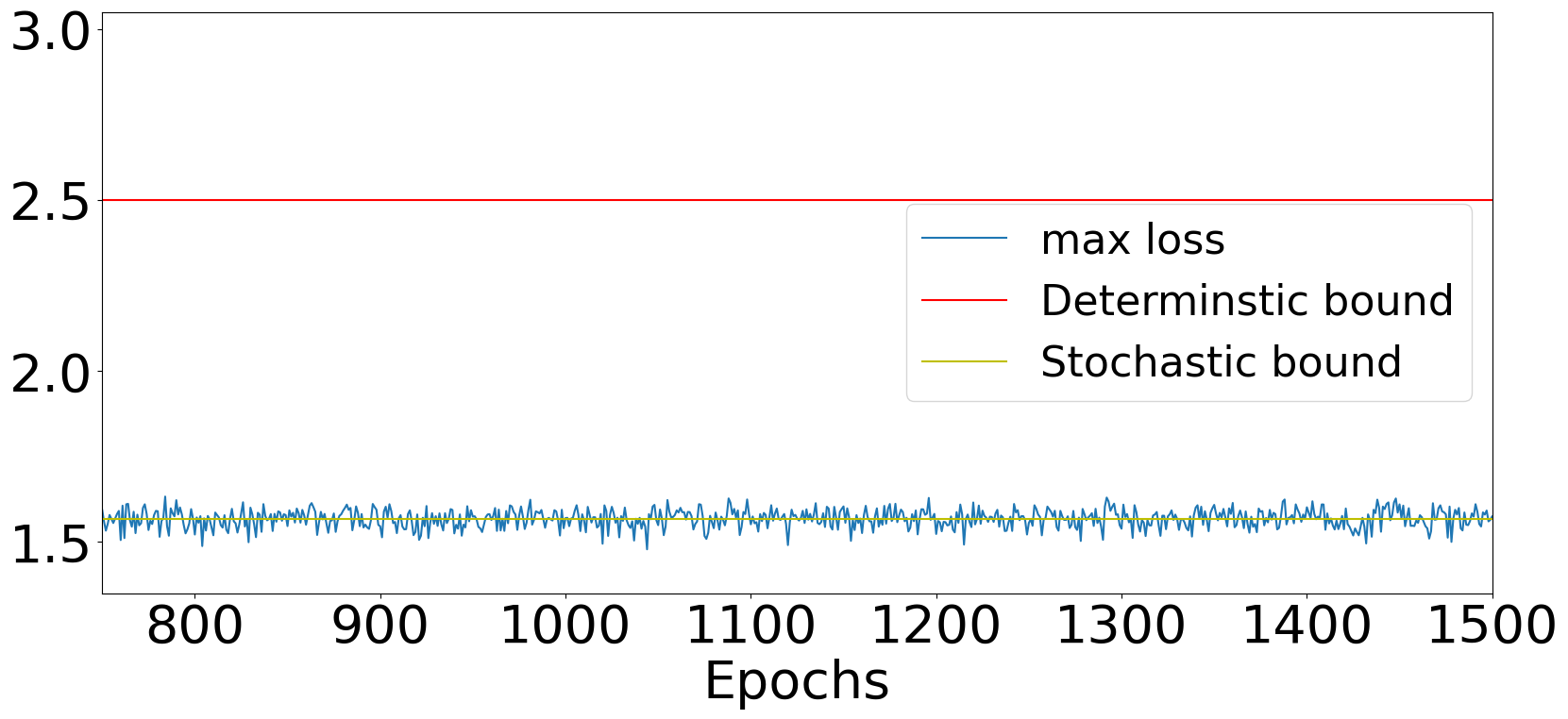}
    	\caption{Results with $\t=0.01$ (left panel) and $\t=0.0348$ (right panel), decreasing the value of $\t$ leads to worse bound and worse convergence}
    	\label{fig:avg2}
    \end{figure}

The results with $B_r=B_s=0.1$ are shown in Figure~\ref{fig:avg1} and Figure~\ref{fig:avg2}. The loss trajectory (blue line) observed with the value suggested by Corollary~\ref{corollary2}, $\t=0.0348$, is the trajectory that has the lowest level . Our theorems correctly predict that decreasing the value of $\t$ is sometimes not beneficial in terms of convergence, see Figure~\ref{fig:avg2}.

\subsection{Experimental results on original MNIST}
In this section the results obtained on the original MNIST dataset are reported. In contrast to the experiments in the main body of the manuscript, PCA is not used to reduce the size of the inputs.

\begin{figure}[!ht]
\centering

\begin{minipage}[t]{0.47\textwidth}
\centering
  \includegraphics[width=0.98\linewidth]{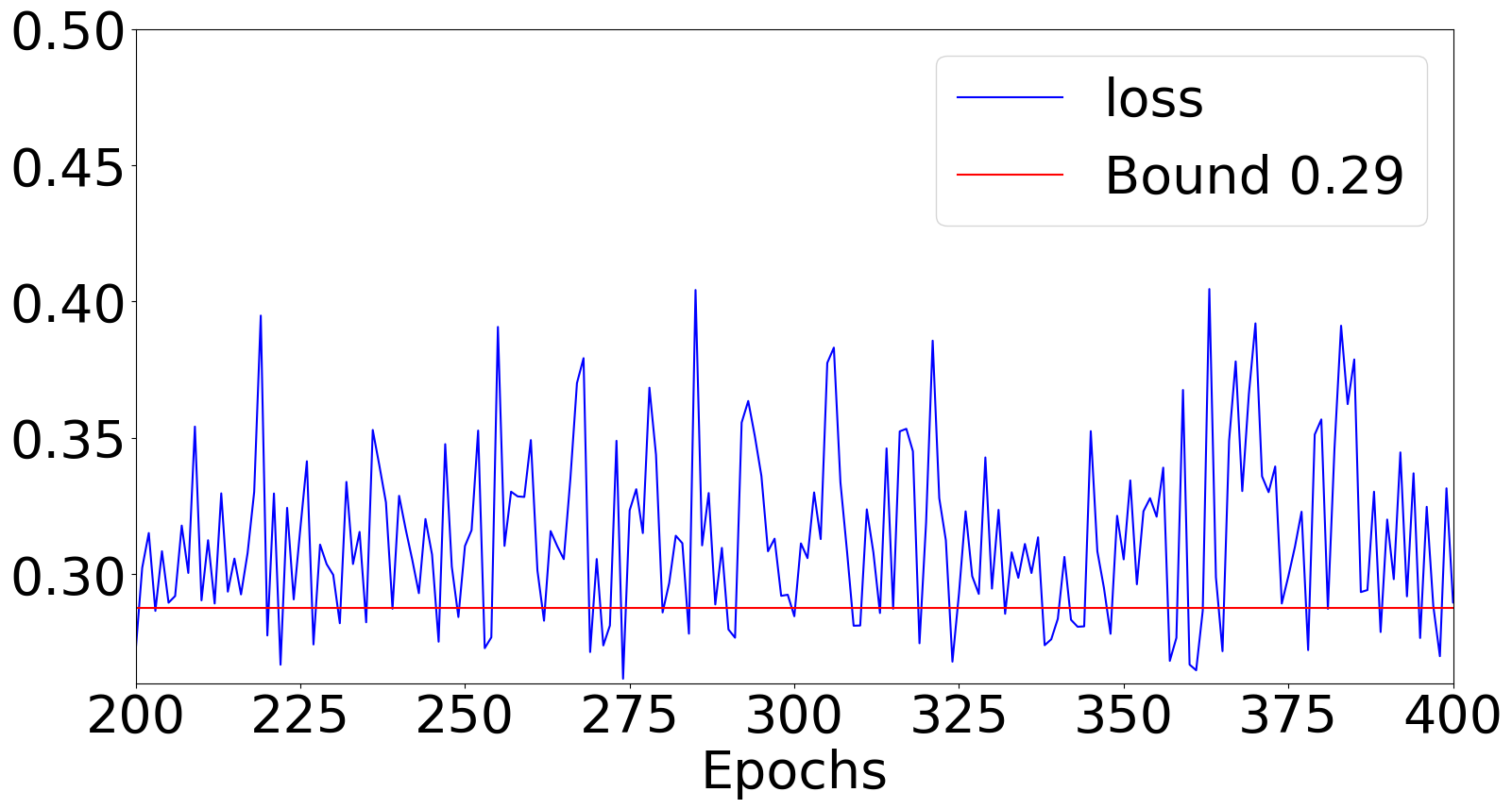}
    	\caption{Logistic regression trained using single-precision SGD and a fixed learning rate.}
\label{fig:fullmnistbase}
\end{minipage}%
\hspace{.2 in}
\begin{minipage}[t]{0.47\textwidth}
\centering
     \includegraphics[width=0.98\linewidth]{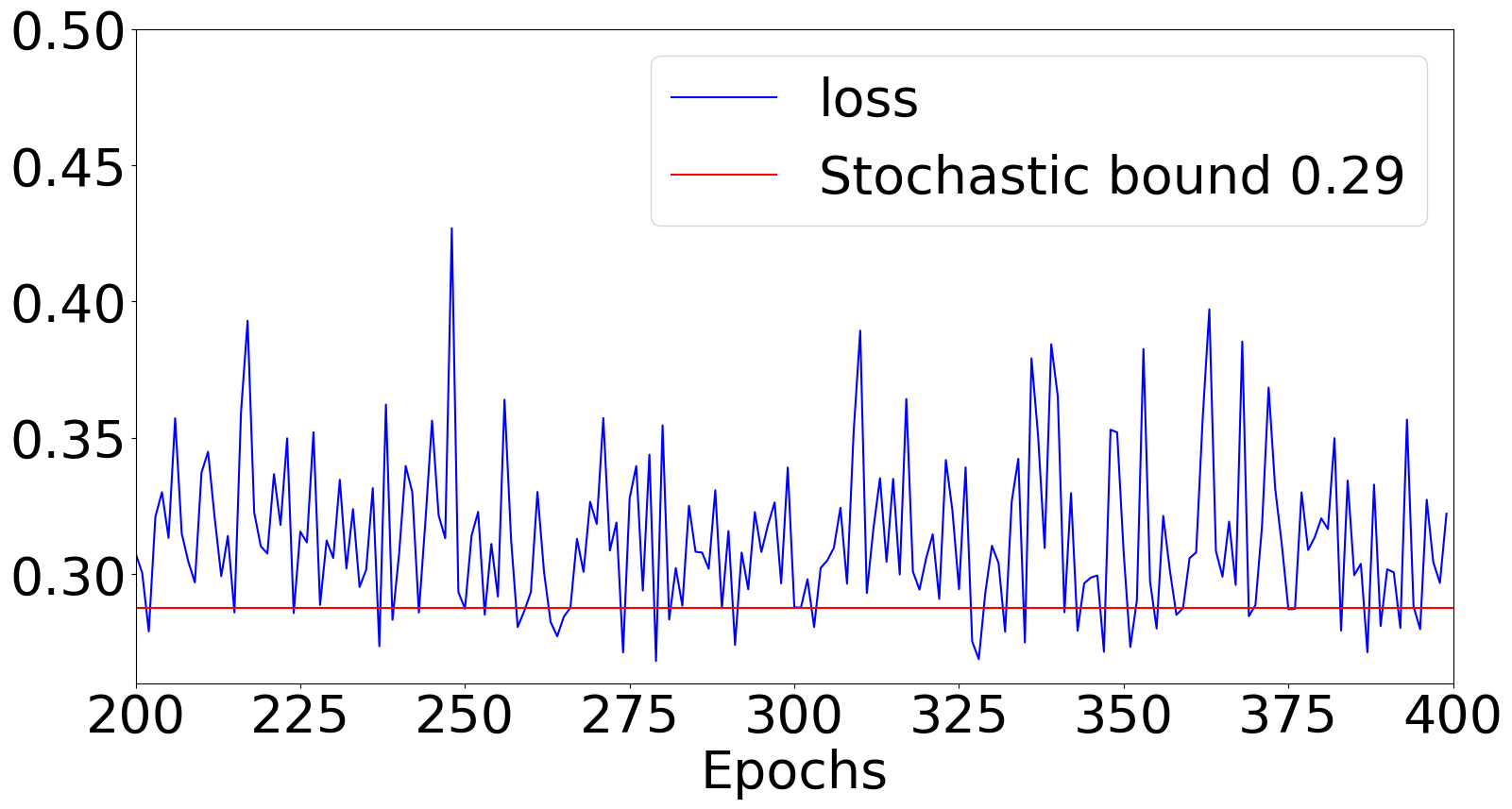}
    	\caption{Logistic regression trained using Bfloat gradients with accumulator size of 50, and single-precision weight update ($\s_k = 0$ and $\r_k\neq 0$).}
    	\label{fig:fullmnisquant}
\label{fig:prob1_6_1}
\end{minipage}

\end{figure}

 Figure \ref{fig:fullmnistbase} demonstrates that the loss trajectory (blue line) has a limit point in the proximity of the optimal point of the convex loss function.
 Figure \ref{fig:fullmnisquant} shows the loss trajectory when the weight update is in single precision and only gradient computations are performed using Bfloat number format. In this experiment, the stochastic bound is numerically equal to the single-precision SGD,  confirming what already observed in Figure~\ref{fig7}.

\section{More motivations for the quasi-convex hypothesis}
\begin{figure}[!ht]
        \centering
        % \begin{center}
        \hspace{-.3 in}
        \includegraphics[width=0.465\linewidth]{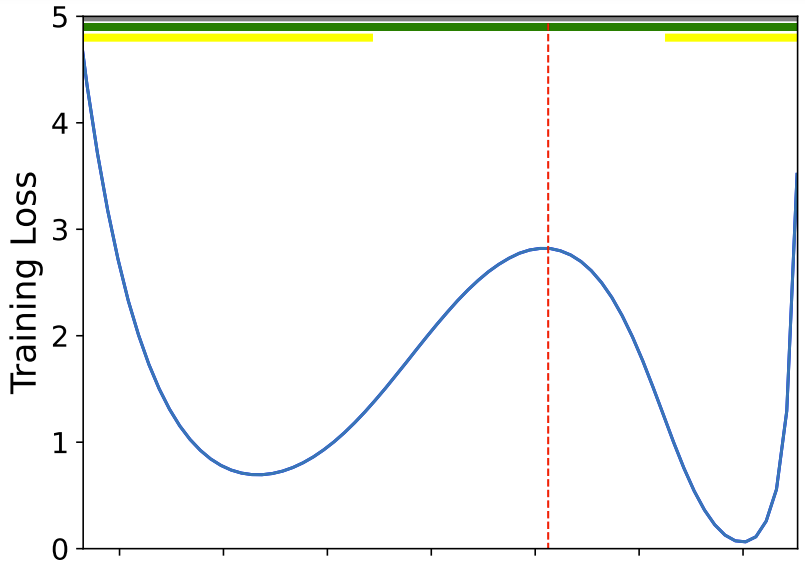}\hspace{.5 in}\includegraphics[height=0.325\linewidth]{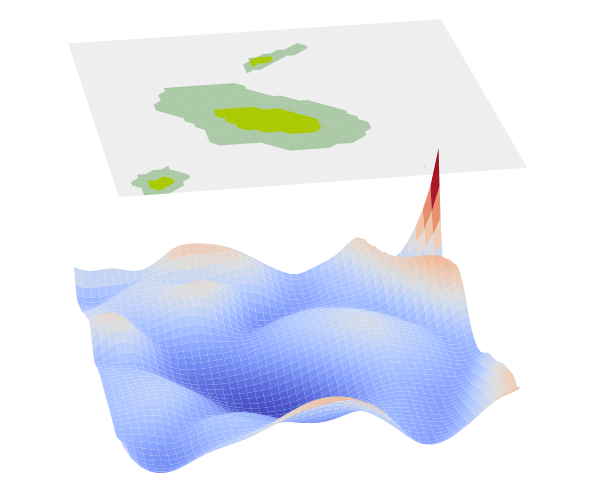}
    	\caption{The quasi-convex regions (in green) are larger than the convex regions (in yellow).}
    % 	\end{center}
    	\label{fig:qcrnet56}
    \end{figure}

To further motivate the quasi convex hypothesis we show the ResNet-56 loss-landscape projection in two and three dimensions without skip connections over the CIFAR10 dataset, see Figure~\ref{fig:qcrnet56}. In this figure, the convex regions and the quasi-convex regions are highlighted. The quasi-convex regions are larger than the convex regions. This means that our theory is applicable in a larger domain of the loss function.

\end{document}



% Supplementary material: To improve readability, you must use a single-column format for the supplementary material.
\onecolumn
\aistatstitle{Supplementary Materials}

\section{Detailed proofs}

\subsection{Theorem 1}
Suppose that $\exists \delta>0 \; s.t. \, f(x_k)>f^*+L(Rd+\frac{t}{2}(1+R)^2+\delta)^p+\epsilon\quad \forall\, k>\bar{k}$, using the Holder condition we now that, given $x^*\in X^*$,  $\forall \, x\in \BB(x^*,Rd+\frac{t}{2}(1+R)^2+\delta)$ it holds, $\forall\, k>\bar{k}$, 
\[
f(x)-f^*\leq L \dist(x,X^*)^p\leq L(Rd+\frac{t}{2}(1+R)^2+\delta)^p\leq f(x_k)-f^*-\epsilon
\]
This means that $\BB(x^*,Rd+\frac{t}{2}(1+R)^2+\delta)\subset  \S_{f,f(x_k)-
\epsilon}\subset  \S_{f,f(x_k)} \quad \forall\, k>\bar{k}.$

We can use this to estimate
    \[
\|x_k-\bx\|^2-2t({\color{red}\langle x_k-\bx,g_k\rangle+\langle x_k-\bx,r_k\rangle-\frac{t}{2}\|\frac{g_k}{\|g_k\|}+r_k\|^2})\leq
\]
\[
\|x_k-\bx\|^2-2t({\color{red}-\frac{t}{2}(1+R^2)-R(t^2+d^2-2tP)^{\frac{1}{2}}+P})
\]
And now we can compute the minimum value for $P$ to complete the proof:

\begin{align}
    \begin{split}
      -\frac{t}{2}(1+R^2)-R(t^2+d^2-2tP)^{\frac{1}{2}}+P>0 \\
       \frac{t^2}{4}(1+R^2)^2+P^2-Pt(1+R^2)>R^2(t^2+d^2-2tP)\\
       P^2+Pt(2R^2-(1+R^2))+\frac{t^2}{4}(1+2R^2+R^4)-R^2t^2-R^2d^2>0\\
        P^2+Pt(R^2-1)+\frac{t^2}{4}(1-2R^2+R^4)-R^2d^2>0\\
        P^2+Pt(R^2-1)+\frac{t^2}{4}(R^2-1)^2-R^2d^2>0\\
        P>\frac{t(1-R^2)+\sqrt{t^2(R^2-1)^2-t^2(R^2-1)^2+4R^2d^2}}{2}\\
        P>\frac{t(1-R^2)+2Rd}{2}=\frac{t(1-R^2)}{2}+Rd
    \end{split}
\end{align}
Notice this only make sense if from the beginning $P<\frac{t^2+d^2}{2t}$ that is implied by $d>P$ and we also need $P>\frac{t}{2}(1+R^2)$, so we need to check that $d>\frac{t}{2}(1+R^2)$ otherwise there is no solution (this basically means that the basic bound provided by $P=d$ is the best we can get).
 So we can use $$P=\min\{\max\{\frac{t}{2}(1+R^2),\frac{t(1-R^2)}{2}+Rd\},d\}$$
By the way, in real instances, the bound provided by $P=d$ will never be meaningless, so we should avoid that case.

Simply form the definition of the algorithm we have that, $\forall \bx\in X$,

\[
\|x_{k+1}-\bx\|^2\leq\|x_k-t\hat{g}_k-\bx\|^2=\|x_k-\bx\|^2-2t\langle x_k-\bx,\frac{g_k}{\|g_k\|}\rangle+2t\|x_k-\bx\|\|r_k\|+t^2\|\frac{g_k}{\|g_k\|}+r_k\|^2\leq
\]

\[
\|x_k-\bx\|^2-2t(\langle x_k-\bx,\frac{g_k}{\|g_k\|}\rangle-Rd-\frac{t}{2}(1+R)^2)
\]
using \REF{th3eq1} we obtain
\[
\|x_{k+1}-x^*\|^2\leq\|x_k-x^*\|-2t\delta\leq \dots\leq \|x_0-x^*\|-2t(k-\bar{k}+1)\delta\quad \forall\, k>\bar{k}
\]
that lead to contraddiction for $k$ big enough.

\subsubsection{Lemma \ref{lemma_aux1}}
Lets give a bit of context:
in the proof we derive first that
\[
\langle x_k-\bx,\frac{g_k}{\|g_k\|}\rangle>P+\delta>0
\], both $P>0$ and $\delta>0$, 
and then we use that in the following estimate
\[
\|x_{k+1}-\bx\|^2\leq\|x_k-t\hat{g}_k-\bx\|^2=\|x_k-\bx\|^2-2t\langle x_k-\bx,\frac{g_k}{\|g_k\|}\rangle-2t\langle x_k-\bx,r_k\rangle+t^2\|\frac{g_k}{\|g_k\|}+r_k\|^2\leq
\]

\[
\|x_k-\bx\|^2-2t({\color{red}P+\langle x_k-\bx,r_k\rangle-\frac{t}{2}\|\frac{g_k}{\|g_k\|}+r_k\|^2})
\]
where we need the red part to be estimated by $\delta$ and we can conclude. The final bound will highly depend on $P$ (greater $P$ implies greater bounds).

In the original proof, we derive

\[
\|x_k-\bx\|^2-2t({\color{red}P+\langle x_k-\bx,r_k\rangle-\frac{t}{2}\|\frac{g_k}{\|g_k\|}+r_k\|^2})\leq
\]
\[
\leq \|x_k-\bx\|^2-2t({\color{red}P-Rd-\frac{t}{2}(1+R)^2})
\]
and then choose $P=Rd+\frac{t}{2}(1+R)^2$.

In particular, we are using
\[\langle x_k-\bx,r_k\rangle\geq -\|x_k-\bx\|\|r_k\|\]
and
\[\|\frac{g_k}{\|g_k\|}+r_k\|\leq (\|\frac{g_k}{\|g_k\|}\|+\|r_k\|)\]
These two last inequalities can be an equalities only if $r_k = -\alpha x_k-\bx$ and $r_k = \beta g_k$ with $\alpha,\;\beta>0$, but this would imply $\langle g_k, x_k-\bx\rangle = \langle r_k/\beta, -r_k/  \alpha\rangle=-\frac{1}{\alpha\beta}\|r_k\|^2<0$, so when $g_k \in \partial^* f(x_k)$ and $\bx \in \S_{f,f(x_k)}$ this is clearly impossible for the definition of a quasi-sub-differential.

\begin{figure}[H]
    \centering
  \begin{tikzpicture}[scale=1,inner sep=0pt, outer sep=2pt,>=latex]
\draw[->](0,0)--(1,0.2)node[midway, above]{$g$}; 
\draw[->](0,0)--(0,3)node[midway,right]{$x^*-x$};
\draw[->](0,0)--(0.15,-1)node[right]{$r$};
\draw(0,0)node{•}node[below,xshift=-5]{$x$};
\draw[dashed, color=red](-3.5,0)--(3.5,0);
\draw(0,3)node[right]{$x^*$};
\draw[draw=red,fill=red, opacity=0.1] (-3.5,0) rectangle (3.5,3);
\end{tikzpicture}
    \caption{Given $x$ and $x^*$, $g$ needs to have positive scalar product with $x^*-x$ (red zone). So it s not possible for $r$ to be opposite to $x^*-x$ and colinear with $g$ at the same time}
    \label{fig:my_label}
\end{figure}

\begin{figure}[H]
    \centering
  \begin{tikzpicture}[scale=1,inner sep=0pt, outer sep=2pt,>=latex]
\draw[->](0,0)--(1,0.2)node[midway, above]{$tg$}; 
\draw[->, color=purple](1,0.2)--(0,3)node[right, midway]{$x^*-(x+tg)$}; 
\draw[->, color=blue, dashed](1,0.2)--(1.5,-1.2)node[right, midway, color=blue]{$r$};
\draw[->](0,0)--(0,3)node[midway,left]{$x^*-x$};
\draw[->,color=blue](0,0)--(0.5,-1.4)node[right]{$r$};
\draw(0,0)node{•}node[below,xshift=-5]{$x$};
\draw[dashed, color=red](-3.5,0)--(3.5,0);
\draw(0,3)node[right]{$x^*$};
\draw[draw=red,fill=red, opacity=0.1] (-3.5,0) rectangle (3.5,3);
\end{tikzpicture}
    \caption{Given $x$, $x^*$ and $g$ the worst case seems to be $-r=x^*-(x+tg)$ (rescaled to have maximum norm) }
    \label{fig:my_label}
\end{figure}

\begin{lemma}
\label{lemma_aux1}
Consider the following optimization problem:
\begin{align*}
    F^*=\max_{x,r\in\RR^n}F(x,r)=\max_{x,r\in\RR^n}\,F(x,r)=\;& t\|g+r\|^2-2\langle x,g+r\rangle\\
    & \langle g,x\rangle \geq P &(\lambda)\\
    & \|r\|^2\leq R^2 &(\gamma)\\
    & \|x\|^2\leq d^2 &(\xi)
\end{align*}
where $\|g\|=1$, $R<1$, $d\geq P>0$ and $t>0$.
Then $F^*\leq  t(1+R^2)+2R(t^2+d^2-2tP)^{\frac{1}{2}}-2P $
\end{lemma}

\subsection{Lemma \ref{lemma_aux1}}

The gradient of the objective function is 
\[
\nabla F(\cdot)= (-2(g+r),2t(g+r)-2x)
\]
The objective is convex so the optimum is not reached in any internal point. 
Using lagrangian multipliers, from the condition on the gradient component with respect to $r$, we obtain
\begin{align*}
    \nabla_r F = \lambda \nabla_r (\|r\|^2-R^2)\\
    2t(g+r)-2x = \lambda 2r\\
    r = \frac{x-t g}{t-\lambda}
\end{align*}
where we can exclude $t=\lambda$ since that implies $x=tg$ and in that case the objective function is just $ t+t\|r\|^2-2t<0$ with optimal solution $\|r\|=R$.
Now lets rewrite the  objective function as
\[
t+2\langle tg-x,r\rangle+t\|r\|^2-2\langle x,g\rangle
\]

and denote with $r^*$ and $x^*$ the optimal values of $r$ and $x$.

We now prove by contradiction some properties of the optimal solutions that will allow us to compute the optimal value:
\begin{itemize}
\item Assume $\langle r^*,tg-x\rangle<0$, then consider $\tilde r =-r^*$, we have $\langle tg-x,\tilde r\rangle>0>\langle tg-x,r^*\rangle $, so $\tilde r$ will have better objective value than $r^*$,  this would violate the optimality of $r^*$, so we have $\langle r^*,tg-x\rangle\geq0$;
    \item Assume $\|r^*\|<R$, then consider $\tilde r= R\frac{r^*}{\|r^*\|}$, we have $\langle tg-x,\tilde r\rangle-\langle tg-x,r^*\rangle=(\frac{R}{\|r^*\|}-1)\langle tg-x,r^*\rangle\geq0$ due also to the first point, since we also have $\|\tilde r\|>\|r^*\|$  this would violate the optimality of $r^*$, so we have  {\color{red}$\|r\|=R$};
    \item assume $\langle x^*,g\rangle>P$, consider $\tilde x=x^*-(\langle x^*,g\rangle-P)g$, then
    $F(x^*,r^*)-F(\tilde x,r^*)=2\langle \tilde x-x^*,g+r\rangle=-2\langle (\langle x^*,g\rangle-P)g,g+r\rangle=-2(\langle x^*,g\rangle-P)-2\langle (\langle x^*,g\rangle-P)g,r\rangle\leq -2(\langle x^*,g\rangle-P)(1-\|g\|\|r\|)<0$, for the same reason of previous points this is not possible and we have:
    {\color{red}$\langle x,g\rangle=P$}
    \item The previous two points and the the formula for $r$ that we found using lagrangian multipliers imply that $r^*=\alpha (tg-x)$ with $\alpha=\frac{R}{\|tg-x\|}$
    \item Assume $\|x^*\|<d$, then consider $y$ s.t. $\langle g,y\rangle=0$ and $\langle x^*,y\rangle\geq0$, consider the point $\tilde x= x^*+a y$ with $a>0$ chosen such that $\|\tilde x\| = d$ (we can suppose also $\|y\|=1$). If $g$ and $x^*$ are not colinear I can choose $y$ such that $\langle x^*,y\rangle>0$, then
    $F(x^*,r^*)-F(\tilde x, r^*)=2\langle \tilde x-x^*,g+r^*\rangle=2\langle a y,g+r^*\rangle=2\langle a y,r^*\rangle=2\langle a y,\alpha (tg-x^*)\rangle=-2a\alpha\langle y,x^*\rangle<0$, where we used the third point.
    
    If $x=Pg$, then $\langle x^*,y\rangle=0$, we consider $\tilde r=\frac{R}{\|tg-\tilde x\|}(tg-\tilde x)=\beta (t-P)g-ay$ ($\beta=\frac{R}{\|tg-\tilde x\|}$) but then:
    $F(x^*,r^*)-F(\tilde x,\tilde r)=2(\langle(t-P)g,\sgn(t-P)Rg \rangle-\langle(t-P)g-ay,\beta (t-P)g-ay  \rangle)=2(\langle\sgn(t-P)(t-P)g,Rg \rangle-\beta\langle(t-P)g-ay, (t-P)g-ay  \rangle)=2(|t-P|R-\beta((t-P)^2+a^2))=2(|t-P|R-R\sqrt{(t-P)^2+a^2})=R\sqrt{(t-P)^2+a^2}(\frac{|t-P|}{\sqrt{(t-P)^2+a^2}}-1)<0$.
    
    In both cases we find a point with better objective value than $(x^*,r^*)$, this is not possible and so we get {\color{red}$\|x^*\|=d$ (this last point hold only in a space with at least 2 dimensions but we can still estimate the norm of $x^*$ with $d$, of course in that case the value of P might not be optimal)}
    \item all together this points imply
    \begin{equation}
        F(x^*,r^*)= t(1+R^2)+2R(t^2+d^2-2tP)^{\frac{1}{2}}-2P
    \end{equation}
\end{itemize}        
\subsection{Lemma 4.1}
In the case where there is an error in the summation, we get:
\begin{align}
    \begin{split}
        \max\{\frac{t}{2}(1+R^2+\frac{S^2}{t^2}+\frac{2SR}{t}),\frac{t(1-R^2-\frac{S^2}{t^2}-\frac{2RS}{t})}{2}+Rd+\frac{Sd}{t}\}=\\=
        \max\{\frac{t}{2}(1+R^2)+\frac{S^2}{2t}+SR,\frac{t}{2}(1-R^2)-\frac{S^2}{2t}-RS+Rd+\frac{Sd}{t}\}
    \end{split}
\end{align}
Studying this function we get 
\begin{align}
    \begin{cases}
    \frac{t}{2}(1+R^2)+\frac{S^2}{2t}+SR & \text{if }t\geq \frac{d-S}{R}\\
    \frac{t}{2}(1-R^2)-\frac{S^2}{2t}-RS+Rd+\frac{Sd}{t} &\text{otherwise}
    \end{cases}
\end{align}
Studying each part we get that the minimum of the first part is reached in $t_1=\frac{S}{\sqrt{1+R^2}}$
and the minimum of the second part is reached in $t_2=\sqrt{\frac{S(d-2S)}{1-R^2}}$
So the global minimum is one of this 3 points $\{t_1.t_2,\frac{d-S}{R}\}$.
\section{Details of the experiments setting}

\section{More on floating points}
\section{More motivations for the quasi-convex hypothesis}

\section{ANNs bound computation}

\section{Additional experiments}

\vfill